\documentclass[lettersize,journal]{IEEEtran}

\usepackage{graphics}
\usepackage{amsmath,amsfonts}
\usepackage{algorithm}
\usepackage[noend]{algpseudocode}
\usepackage{array}
\usepackage{textcomp}
\usepackage{stfloats}
\usepackage{url}
\usepackage{verbatim}
\usepackage{cite}
\usepackage{xcolor}
\usepackage{todonotes}
\usepackage{float}
\usepackage{graphicx}
\usepackage{subcaption}
\usepackage{multirow}
\usepackage{mathtools}
\usepackage{bbm}
\usepackage{environ}
\usepackage{physics}
\usepackage{booktabs}

\DeclareUnicodeCharacter{221E}{}

\def\BibTeX{{\rm B\kern-.05em{\sc i\kern-.025em b}\kern-.08em
    T\kern-.1667em\lower.7ex\hbox{E}\kern-.125emX}}
\usepackage{balance}

\title{A Hamilton-Jacobi Reachability-Guided Search Framework for Efficient and Safe Indoor Planar Robot Navigation}

\author{Hanyang Hu, Cameron Siu, Mo Chen%
\thanks{This work has been submitted to the IEEE for possible publication. Copyright may be transferred without notice, after which this version may no longer be accessible.}%
\thanks{$^{1}$Hanyang Hu, Cameron Siu, and Mo Chen are with Simon Fraser University, Burnaby, BC V5A 1S6, Canada.}%
}

\begin{document}

\maketitle

\begin{abstract}
  Autonomous navigation requires planning to reach a goal safely and efficiently in complex and potentially dynamic environments.
  Graph search-based algorithms are widely adopted due to their generality and theoretical guarantees when equipped with admissible heuristics.
  However, the computational complexity of graph search grows rapidly with the dimensionality of the search space, often making real-time planning in dynamic environments intractable.
  In this paper, we combine offline Hamilton-Jacobi (HJ) reachability with online graph search to leverage the complementary strengths of both.
  Precomputed HJ value functions, used as informative heuristics and proactive safety constraints, amortize online computation of the graph search process.
  At the same time, graph search enables reachability-based reasoning to be incorporated into online planning, overcoming the long-standing challenge of HJ reachability requiring full knowledge of the environment.
  Extensive simulation studies and real-world experiments demonstrate that the proposed approach consistently outperforms baseline methods in terms of planning efficiency and navigation safety, in environments with and without human presence.
\end{abstract}


\section{Introduction}
\label{intro}

Indoor planar robot navigation requires generating trajectories that are both computationally efficient and safe in environments populated with static obstacles and moving agents.
Achieving these objectives simultaneously remains challenging, particularly when real-time performance and dynamic feasibility are required.

A broad range of planning paradigms has been proposed to address this problem.
\textbf{Sampling-based methods}, such as rapidly-exploring random tree (RRT) \cite{lavalle1998rapidly} and probabilistic roadmap (PRM) \cite{kavraki1996probabilistic}, construct collision-free paths by random exploration and are probabilistically complete \cite{orthey2023sampling}.
Their asymptotically optimal extensions, including RRT* and PRM* \cite{karaman2011sampling} and fast marching tree (FMT*) \cite{janson2015fast}, further improve solution quality as the number of samples increases.
However, these methods lack deterministic performance guarantees within finite time, and often require post-processing to ensure trajectory smoothness.
\textbf{Optimization-based methods}, including model predictive control (MPC), directly incorporate dynamics and constraints to generate smooth and dynamically feasible trajectories \cite{schwenzer2021review}.
Their performance, however, depends heavily on initialization and may degrade in cluttered environments due to local minima.
\textbf{Learning-enhanced methods} leverage data-driven policies for goal reaching and obstacle avoidance \cite{zhu2021deep}, but their safety, interpretability, and generalization in unseen environments remain open concerns for safety-critical tasks.

\textbf{Graph search-based methods} provide an appealing alternative due to their interpretability and theoretical guarantees when equipped with admissible heuristics \cite{korf2000recent}.
By discretizing the state space and searching over a graph representation, algorithms such as A* \cite{hart1968formal} and D* \cite{stentz1994optimal} can guarantee optimality under suitable conditions.
Nevertheless, their computational complexity grows rapidly with the dimensionality of the state space.
Despite the use of anytime algorithms \cite{van2011anytime, hansen2007anytime}, uninformed search may expand a large number of nodes that ultimately correspond to unsafe or dynamically infeasible states in dynamic or safety-critical environments.
Consequently, the effectiveness of search-based planning critically depends on the design of informative heuristics and predictive pruning strategies that can reduce unnecessary exploration while preserving safety constraints.

HJ reachability is a robust optimal control method that computes the optimal control under worst-case adversarial disturbances.
Solutions to Hamilton-Jacobi-Isaacs (HJI) partial differential equations (PDEs) yield value functions that characterize reachable tubes or minimal arrival times \cite{bansal2017hamilton}.
In particular, backward reachable tubes (BRTs) identify states that can be driven to a target or failure set within a given horizon despite bounded disturbances \cite{chen2018hamilton}, while time-to-reach (TTR) functions quantify the minimum time required to reach the target \cite{yang2013one}. 
These value functions encode useful information about dynamic feasibility and safety from the system dynamics.

Despite their strong theoretical guarantees, directly deploying HJ value functions for online planning remains challenging in practical robotic navigation tasks.
Value functions are typically precomputed offline with respect to specific targets or environmental configurations, making them difficult to adapt to changing goals or partially unknown environments.
Moreover, using BRTs purely as hard safety filters may conflict with goal-directed planners, potentially leading to deadlock or oscillatory behaviors.
These observations suggest that HJ reachability is more effective as a source of principled guidance rather than as a standalone planner or safety mechanism.

\begin{figure}[htb]
    \centering
    \includegraphics[width=0.48\textwidth]{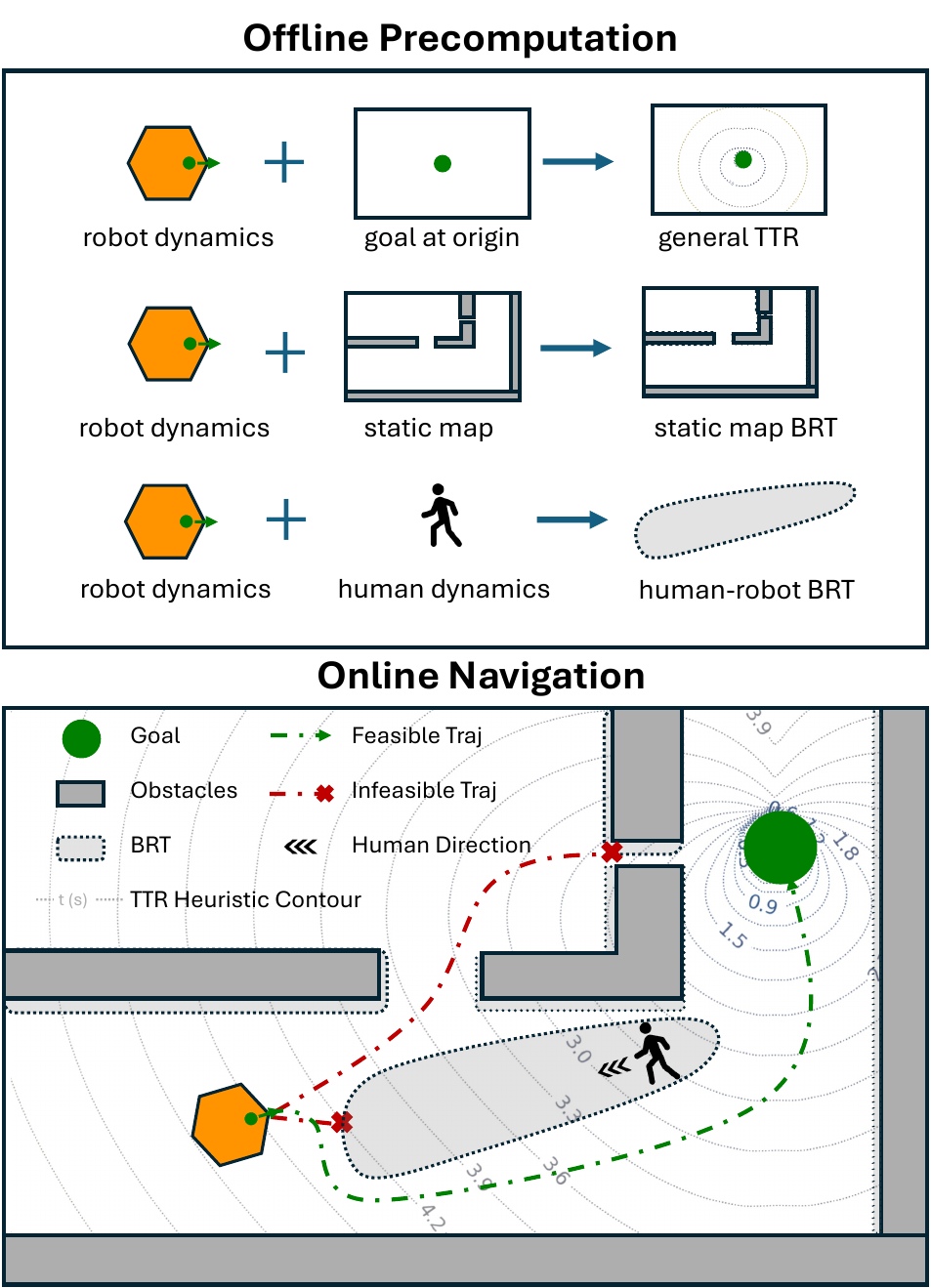}
    \caption{Overview of the proposed HJ reachability-guided search framework. In the offline stage, value functions representing a general TTR, a static-map BRT, and a human-robot BRT are computed. Online, these value functions are integrated into the graph search algorithm to improve search efficiency and guarantee safety.}
    \label{overview}
\end{figure}

In this paper, we propose an HJ reachability-guided search framework for indoor planar robot navigation that integrates offline reachability analysis with online graph search.
We embed precomputed HJ value functions into the search process to enhance both efficiency and safety awareness without requiring reachability recomputation during online planning, as illustrated in Fig.~\ref{overview}.
Specifically, we precompute three HJ-reachability value functions: (1) a generall TTR function that characterizes dynamically informed heuristic guidance, (2) a static-map avoid value representing a BRT that encodes collision avoidance with known static obstacles, and (3) a task-specific human-robot value representing a BRT that captures collision avoidance with a controlled human agent.
During online planning, TTR replaces conventional distance-based heuristics to fasten convergence toward the goal, while the BRTs proactively prune states that will lead to static or dynamic collisions.
By combining the structure of graph search with the predictive safety characterization of HJ reachability, our proposed framework amortizes the online search process, and achieves improved search efficiency and enhanced safety awareness in dynamic indoor environments.
Extensive simulation studies and real-world experiments in human-populated settings validate the effectiveness of the proposed approach.
The main contributions of this paper are summarized as follows:
\begin{enumerate}
    \item We propose an HJ reachability-guided search framework that integrates precomputed reachability value functions into graph search to jointly improve planning efficiency and safety.
    \item We introduce a unified integration strategy that leverages TTR for heuristic acceleration and BRTs for predictive pruning of static and dynamic hazards.
    \item We demonstrate the practical effectiveness of the proposed framework through high-fidelity simulations and real-world experiments in environments with moving humans.
\end{enumerate}

\section{Related Work}
\subsection{Heuristic Design and Safety-Aware Graph Search-Based Planning}
Graph search-based planners, such as A* \cite{hart1968formal} and its variants, rely heavily on the quality of heuristic functions to guide the search process toward the goal area \cite{korf2000recent}.
When equipped with admissible heuristics, these methods provide optimality guarantees \cite{korf2000recent}, while weighted and anytime extensions improve real-time performance by trading optimality for search efficiency \cite{pohl1970heuristic, likhachev2005anytime, van2011anytime}.
More informative heuristics can substantially reduce unnecessary node expansions and thereby improve search efficiency.

To address dynamical feasibility, kinodynamic search methods and motion primitive-based planners incorporate system dynamics into the graph construction \cite{plaku2007discrete, pivtoraiko2011kinodynamic}. 
These approaches ensure dynamically feasible motions by constructing edges based on system dynamics, and have been successfully applied to nonholonomic systems, like Dubins Car \cite{macenski2024open}. 
However, the associated heuristics are typically based on geometric distance metrics or simplified cost-to-go approximations that do not fully capture worst-case disturbance effects or predictive safety constraints.

Recently, learning-based techniques have been explored to accelerate heuristic evaluation or approximate value functions for search \cite{savinov2018semi, guez2018learning, ferber2022neural, yonetani2021path}. 
While these methods can significantly reduce node expansions in structured environments, their performance highly depends on training data and does not generally provide formal safety guarantees.

Safety-aware search strategies have also been studied, including constraint-based pruning \cite{parthasarathy2023c, jamgochian2024constrained, parimi2025risk} and risk-aware cost shaping \cite{primatesta2019risk, yan2023safe, macenski2024open}. 
These methods aim to reduce the exploration of unsafe nodes by incorporating obstacle proximity or collision penalties into the cost function. 
Nevertheless, such mechanisms are often reactive and geometry-based, lacking model-based worst-case reasoning about the system dynamics.

In summary, prior work has improved search efficiency through heuristic refinement, motion primitive design, or learned priors. 
However, few approaches integrate control-theoretic safety analysis directly into the heuristic evaluation and node expansion process. 
This observation motivates the incorporation of reachability-based value functions into graph search as a principled mechanism for both heuristic guidance and predictive pruning.

\subsection{HJ Reachability-Based Control and Planning}
\label{sec:related-HJ}
HJ reachability provides a rigorous framework for characterizing safety guarantees under bounded disturbances \cite{bansal2017hamilton}. 
The HJI equation and its viscosity solution, the HJ value function, encodes safety and optimality information about the system dynamics.

When the HJ value function is used as a safety filter, a controlled system may use a task-oriented control when the HJ safety value is above some threshold; once the value approaches the safety threshold, the system needs to switch to a safety control computed based on  value function's gradient \cite{borquez2024safety, chen2021fastrack, hsu2023safety, lu2025safe, seo2025uncertainty, nakamura2025generalizing}.

HJ reachability analysis has also been explored for motion planning by directly using value functions to generate trajectories or guide control policies \cite{mitchell2005time, fisac2015reach}.
BRTs are often used as hard safety constraints to prevent the system from entering usnafe regions.
However, when used purely as hard constraints, BRTs may conflict with goal-directed planning objectives, since safety filters can reject intermediate actions that are necessary for eventually reaching the goal.
TTR value functions, on the other hand, can be used more directly for planning.
For example, gradients of TTR can be used to synthesize dynamically feasible motions toward a goal \cite{chen2018robust, agand2022human, lyuttr}. 
However, such approaches typically rely on value functions computed offline for specific targets or environmental configurations, which limits their adaptability to changing goals or partially unknown environments \cite{parkinson2022time}.
In contrast, a general TTR value function enables map-independent guidance, allowing the same value function to remain applicable even when the environment or map changes.
These observations suggest that reachability value functions are often more effective as structured guidance mechanisms than as standalone planning strategies.

Despite their strong theoretical guarantees, HJ reachability value functions have rarely been embedded into graph search algorithms as structural components of heuristic evaluation or predictive pruning.
Most existing approaches treat BRTs as standalone safety certificates or post hoc filters, rather than as mechanisms to shape the exploration process itself.

\section{Preliminaries}

\subsection{Graph Search-Based Methods}
Graph search-based methods model planning problems as graphs, where nodes represent system states and edges correspond to feasible transitions between two sequential states.
These methods aim to find a path from an initial state to a goal state by systematically exploring neighboring nodes based on the predefined evaluation criteria.

One representative method is the A* algorithm.
A* selects the next node to expand by minimizing the sum of the accumulated cost from the initial node and a heuristic estimation of the remaining cost to the goal \cite{hart1968formal}.
When the heuristic is admissible, i.e., the heuristic never overestimates the true cost-to-go from the current node to the goal node, A* algorithm is guaranteed to find the optimal path \cite{edelkamp2011heuristic}.
However, feasible solutions within strict computational time limits are often preferred over optimality in real-time applications.

To address this, the Anytime Nonparametric (ANA*) algorithm finds one suboptimal feasible path rapidly by weighting the heuristic aggressively, then iteratively reduces it to expand the most promising node per iteration, adjusting the greediness of the search as path quality improves \cite{van2011anytime}.
Unlike D Lite* \cite{koenig2002d}, which is well suited for dynamic environments but does not explicitly enforce time constraints, ANA* provides a principled mechanism to trade solution optimality for computation time.
This property aligns well with our problem setting, where the robot must generate a feasible path within a fixed time horizon, prioritizing real-time responsiveness over optimality in partially observable environments. 

\subsection{HJ Reachability Analysis}
Given a general dynamical system subject to disturbances,
\begin{equation}
    \label{dyn:HJ}
    \begin{aligned}
        \frac{d x}{dt} = \dot{x} = f(x, u, d)
    \end{aligned}
\end{equation}
where $x \in \mathbb{R}^n$ is the system state, 
and $u$ and $d$ denote the control input and disturbance, respectively.

\subsubsection{Goal-reaching TTR computation}
Given a target set $\mathcal{T} \subseteq \mathbb{R}^n$, the disturbance-free TTR function is defined as:
\begin{equation}
    \label{OPC-TTR}
    \Phi(x)=\min_{u(\cdot)} \left\{ \tau \ge 0 \mid \xi_x^{u}(\tau) \in \mathcal T \right\}
\end{equation}
where $\Phi(x)$ denotes the TTR value function, and \(\xi_x^{u}(\tau)\) denotes the system trajectory at time \(\tau\) starting from the state \(x\) under the control \(u(\cdot)\).
Here, $\Phi(x)$ represents the minimum time required for the system to reach the target set $\mathcal{T}$ at the current state \(x\) under the optimal control.
This value function $\Phi(x)$ is the viscosity solution to this HJI PDE \cite{yang2013one}:
\begin{equation}
    \label{HJI-TTR}
    \begin{aligned}
        \min _{u \in \mathcal{U}} \left( - \frac{\partial \Phi(x)^{\top}}{\partial x} f(x, u) - 1 \right) &= 0,  x \notin \mathcal{T} \\
        \Phi(x) &= 0, x \in \mathcal{T}
    \end{aligned}
\end{equation}
where the time to reach is zero when the agent is inside of the target set.

\subsubsection{Avoid BRT computation}
Given a failure set \(\mathcal{F} \subseteq \mathbb{R}^n\), represented by an implicit function \(l_{\mathcal{F}}(x)\) such that
\(\mathcal{F} = \{x \in \mathbb{R}^n \mid l_{\mathcal{F}}(x) \le 0\}\), the avoid BRT over the time horizon $[-T, 0]$ is a set of states from which the system is inevitably driven into $\mathcal{F}$ within time $T$.
To characterize this unsafe set, we define the reachability value function:
\begin{equation}
    \label{OPC-BRT}
    V(x,t) = \max_{u(\cdot)} \min_{d(\cdot)} \min_{\tau \in [t,0]} l_{\mathcal{F}}\!\left(\xi^{u,d}_{x,t}(\tau)\right)
\end{equation}
where \(\xi^{u,d}_{x,t}(\tau)\) denotes the system trajectory at time \(\tau\) starting from state \(x\) at time \(t\) under control \(u(\cdot)\) and disturbance \(d(\cdot)\). 
The avoid BRT is then represented as the zero sublevel set of \(V(x,t)\):
\begin{equation}
    \label{HJ-BRT}
    \mathcal{B}(t) = \{ x \in \mathbb{R}^n \mid V(x,t) \leq 0 \}
\end{equation}
Here, the value function $V(x, t)$ is the viscosity solution to the following HJI PDE within time $[-T, 0]$ \cite{crandall1983viscosity}:
\begin{equation}
    \label{HJI-BRT}
    \begin{aligned}
         & \min \left\{\frac{\partial V}{\partial t} + H \left(x, p\right), l_{\mathcal{F}}(x)-V(x, t) \right\}
        =0, \\
         & \quad t \in[-T, 0]
    \end{aligned}
\end{equation}
where the optimal Hamiltonian $H$ is calculated as:
\begin{equation*}
    H(x, p)= \max _{u \in \mathcal{U}} \min_{d \in \mathcal{D}}  p^{\top} f(x, u, d)
\end{equation*}
with $p = \frac{\partial V}{\partial x}$.
As the time horizon \(T\) increases, the avoid BRT may converge to a time-invariant reachable tube.
In this case, the corresponding value function is independent of time, and the avoid BRT can be denoted as $\mathcal{B} = \{ x \in \mathbb{R}^n | V(x) \leq 0 \}$.

For HJI PDEs \eqref{HJI-TTR} and \eqref{HJI-BRT}, closed-form analytical solutions generally do not exist.
Thus, we employ a Python-based toolbox, OptimizedDP \cite{minh2025optimize}, to compute numerical solutions.

\section{Structural Connections Between Graph Search and HJ Reachability}
\label{sec:graph-hj-connection}

In this section, we formalize the structural relationship between graph search planning and HJ reachability analysis.

Trajectory planning for dynamical systems can be formulated as a continuous optimal control problem (OCP), where the objective is to minimize a cumulative cost subject to system dynamics and safety constraints:
\begin{subequations}\label{COCP}
     \begin{align}
     & \underset{u(\cdot)}{\text{minimize}}
     & & \int_{0}^{T} c(x,u)\,dt  \label{COCP-obj} \\
     & \text{subject to}
     & & \dot{x} = f(x,u),  \label{COCP-dyn} \\
     & & & x \in \mathcal{X}_{\text{safe}},  \label{COCP-safe} \\
     & & & u \in \mathcal{U} \label{COCP-control}
     \end{align}
     \end{subequations}
where $c(\cdot)$ is the running cost function,
and $x(\cdot)$ and $u(\cdot)$ denote the state trajectory and control input, respectively, 
and $\mathcal{X}_{\text{safe}}$ is the safe constraint set.

In general, solving the OCP \eqref{COCP} analytically is often intractable for systems with complex dynamics and constraints. 
A common approximation is therefore to discretize the continuous state space and represent the planning problem as a graph-search problem. 
Let \(G=(\mathcal{V},\mathcal{E})\) denote a directed graph, where each node \(v_i \in \mathcal{V}\) corresponds to a discrete state \(x_i\), and each edge \((v_i,v_j)\in\mathcal{E}\) represents a candidate transition from \(x_i\) to \(x_j\). 
The associated discrete transition cost approximates the running cost accumulated along the corresponding local trajectory segment: $\bar{c}(v_i,v_j) \approx \int_{0}^{\Delta t} c(x(t),u(t))\,dt$.
Then, the graph-search approximation can be written as the following minimum-cost safe path problem:
\begin{subequations}\label{eq:graph-ocp}
\begin{align}
& \underset{(v_0,\dots,v_K)}{\text{minimize}}
& & \sum_{k=0}^{K-1} \bar{c}(v_k,v_{k+1}) \label{eq:graph-ocp-obj} \\
& \text{subject to}
& & v_0 = v_{\mathrm{start}}, \quad v_K \in \mathcal{V}_{\mathrm{goal}}, \label{eq:graph-ocp-startgoal} \\
& & & (v_k,v_{k+1}) \in \mathcal{E}_{\mathrm{feas}}, \quad k=0,\dots,K-1, \label{eq:graph-ocp-dyn} \\
& & & v_k \in \mathcal{V}_{\mathrm{safe}}, \quad k=0,\dots,K. \label{eq:graph-ocp-safe}
\end{align}
\end{subequations}
Here, \(\mathcal{E}_{\mathrm{feas}} \subseteq \mathcal{E}\) denotes the set of dynamically feasible transitions, and \(\mathcal{V}_{\mathrm{safe}} \subseteq \mathcal{V}\) denotes the set of safe nodes. 
Under this approximation, the objective in \eqref{eq:graph-ocp-obj} approximates the cumulative cost in \eqref{COCP-obj}, the feasible-edge constraint in \eqref{eq:graph-ocp-dyn} captures both the system dynamics in \eqref{COCP-dyn} and the admissible control set in \eqref{COCP-control}.
The safe-node constraint in \eqref{eq:graph-ocp-safe} corresponds to the continuous safety constraint in \eqref{COCP-safe}.

One correspondence arises in cost-to-go estimation. 
The solution to the OCP \eqref{COCP} is fundamentally characterized by a value function that gives the optimal remaining cost from any state to the goal. 
In graph search, the heuristic function plays an analogous role by estimating the remaining cost from a node to the goal. 
The TTR function in \eqref{OPC-TTR} is the value function of a minimum-time optimal control problem. 
Therefore, when time is a meaningful proxy for planning cost, the TTR provides a theoretically grounded heuristic for graph search, supplying global, dynamics-aware guidance for the discrete search process.

Another correspondence arises in safety enforcement.
The safety constraint in \eqref{COCP-safe} can be enforced in graph search by pruning unsafe nodes or transitions. 
A more principled pruning mechanism is obtained from the avoid BRT in \eqref{OPC-BRT}--\eqref{HJI-BRT}. 
Since the avoid BRT characterizes the set of states from which the system is inevitably driven into the failure set, nodes whose states lie in the avoid BRT can be pruned before search expansion.
In this sense, the BRT provides a dynamics-aware approximation of the continuous safety constraint in \eqref{COCP-safe}, refining graph search from purely geometric collision checking to proactive safety pruning.

This perspective also extends to interactive settings involving other controlled agents. 
When safety depends not only on the robot state but also on the state of another agent, the continuous safety constraint is defined over an augmented or relative state space. 
The corresponding BRT then captures unsafe agent-robot interactions and can be used to prune graph nodes that would lead to collision under worst-case agent behavior. 
This observation foreshadows the human-robot BRT construction introduced in the next section.

\section{Problem Formulation}
\label{sec:problem}

We consider an indoor planar robot navigation task where the robot aims to reach the goal set $\mathcal{G} \subseteq \mathbb{R}^n$ while avoiding collisions with both static obstacles $\mathcal{O} \subseteq \mathbb{R}^3$ and controlled moving agents, which in this work are humans.
Let $x_\text{rob} \in \mathcal{X}_\text{rob} \subseteq \mathbb{R}^n$ denote the robot state.
The robot dynamics are given by:
\begin{equation}
    \label{dyn:robot}
    \begin{aligned}
        \frac{d x_\text{rob}}{dt} = \dot{x}_\text{rob} = r(x_\text{rob}(t), u_\text{rob}(t))
    \end{aligned}
\end{equation}
where $u_\text{rob}(t)$ constrained by the compact set $\mathcal{U}_\text{rob} \subseteq \mathbb{R}^{n_{u_\text{rob}}}$ is the control,
the system $r: \mathcal{X}_\text{rob} \times \mathcal{U}_\text{rob} \rightarrow \mathcal{X}_\text{rob}$ is Lipschitz continuous to the state given the control policy $u_\text{rob}(\cdot)$.

Similarly, we consider $N$ humans in the environment.
The dynamics of the $i$-th human are given by:
\begin{equation}
    \begin{aligned}
        \frac{d x^i_\text{hum}}{dt} = \dot{x}^i_\text{hum} = h(x^i_\text{hum}(t), u^i_\text{hum}(t)), \ i = 1, \dots, N
    \end{aligned}
\end{equation}
where $x^i_\text{hum} \in \mathcal{X}_\text{hum} \subseteq \mathbb{R}^m$ denotes the state of the $i$-th human,
$u^i_\text{hum}(t) \in \mathcal{U}_\text{hum} \subseteq \mathbb{R}^{n_{u_\text{hum}}}$ represents the corresponding control input.
The function $h: \mathcal{X}_\text{hum} \times \mathcal{U}_\text{hum} \rightarrow \mathcal{X}_\text{hum}$ is assumed to be Lipschitz continuous in the state for admissible control policies $u^i_\text{hum}(\cdot)$.

To formally define collision avoidance, we introduce geometric mapping functions $\psi_\text{rob}: \mathcal{X}_\text{rob} \rightarrow \mathcal{P}(\mathbb{R}^2)$, $\psi_\text{hum}: \mathcal{X}_\text{hum} \rightarrow \mathcal{P}(\mathbb{R}^2)$, and $\psi_\text{obs}: \mathcal{O} \rightarrow \mathcal{P}(\mathbb{R}^2)$.
These mappings associate the state of the robot, humans, and static obstacles with the corresponding regions of the planar workspace $\mathbb{R}^2$ that they physically occupy, where $\mathcal{P}$ denotes the power set.
Under this formulation, the safe planar navigation task is defined as follows: given initial conditions $x_\text{rob}(0)$ and $x^i_\text{hum}(0)$, there exists a finite time $T > 0$ such that the following conditions hold:
\begin{subequations}\label{eq:safe-nav-task}
    \begin{align}
        x_\text{rob}(T) &\in \mathcal{G}, \label{eq:safe-nav-task-goal} \\
        \psi_\text{rob}(x_\text{rob}(t)) \cap \psi_\text{obs}(\mathcal{O}) &= \emptyset,
        \quad \forall t \in [0,T], \label{eq:safe-nav-task-obs} \\
        \psi_\text{rob}(x_\text{rob}(t)) \cap \psi_\text{hum}(x_\text{hum}^i(t)) &= \emptyset,
        \quad \forall t \in [0,T],  \\ 
        \ \forall i \in \{1,\dots,N\}. \label{eq:safe-nav-task-hum}
    \end{align}
    \end{subequations}

\section{HJ Reachability-Guided Navigation}
\label{sec:hj_nav}

In indoor planar robot navigation, a key challenge is to generate feasible, high-quality, and safe trajectories to reach the goal in environments populated with both static obstacles and moving humans.
This task requires not only accurate modeling of the robot dynamics but also efficient search and predictive safety reasoning under real-time constraints.
However, as discussed in Sec.~\ref{intro}, high-fidelity models substantially increase the dimensionality of the search space and can severely degrade search efficiency.
Meanwhile, safety demands more than reactive collision avoidance.
For static obstacles, many planners perform instantaneous collision checking only, lacking predictive capability.
For moving humans, existing methods often depend on online prediction of future human motion to evaluate interaction risk \cite{fisac2018probabilistically, agand2022human}.
In contrast, we adopt a prediction-free online search process and encode both guidance and safety through precomputed HJ reachability value functions.
This design keeps online search process lightweight, since each node expansion involves only state transformations and table look-ups rather than repeated forecasting or online PDE computation.
This offline-online decomposition amortizes the online computational cost of HJ reachability analysis while preserving its predictive power during deployment.

\begin{algorithm}[htb]
    \caption{HJ Reachability-Guided Navigation}
    \label{alg:nav}
    \begin{algorithmic}[1]
        \Statex \textbf{Offline Precomputation}
        \State Compute a general TTR $\Phi(\bar{x}_\text{rob})$;
        \State Compute the static-map BRT value function $V_\text{map}(x_\text{rob})$;
        \State Compute the human-robot BRT value function $V_\text{rel}(x_\text{rel})$.
        \Statex \textbf{Online Navigation}
        \State Initialize robot state $x_\text{rob}$, \textsc{Planner}, and \textsc{Controller}.
        \While{$x_{\text{rob}} \notin \mathcal{G}$}
        \ForAll{detected humans $i \in \mathcal{I}$}
            \State $x_{\text{rel}}^{i} \gets R(\theta)\big(x_{\text{hum}}^{i} - Q x_{\text{rob}}\big)$
        \EndFor
            \State Generate a dynamically feasible trajectory with:
            \State \quad $\Phi(\bar{x}_\text{rob})$ as the guidance heuristic;
            \State \quad $\mathcal{B}_\text{map}$ and $\mathcal{B}_\text{rel}$ as pruning sets.
            \State Execute control: $ x_\text{rob} \gets \textsc{Controller}(\text{trajectory}) $  
        \EndWhile
    \end{algorithmic}
\end{algorithm}

Algorithm~\ref{alg:nav} summarizes how the three precomputed HJ value functions are integrated into the proposed navigation framework.
In the offline precomputation phase (Lines 1--3), three HJ reachability value functions are precomputed and stored for online lookup.
Specifically, $\Phi(\bar{x}_\text{rob})$ is a general goal-centered TTR value function that provides a reusable dynamics-aware estimate of progress toward the goal.
The static-map BRT value function $V_\text{map}(x_\text{rob})$ captures predictive collision risk with respect to the static map.
The human-robot BRT value function $V_\text{rel}(x_\text{rel})$ encodes predictive interaction risk in the robot-centric relative state space.
In Line 4, the robot initializes its state together with the planner and low-level controller.
In Lines 5--7, for each detected human, the planner computes the relative state $x_\text{rel}^{i}$ by transforming the human state into the robot-centric coordinate frame.
This transformation allows the same precomputed human-robot value function to be queried online for different robot poses and different human observations.
In Lines 8--10, the planner expands candidate nodes using $\Phi(\bar{x}_\text{rob})$ as the search heuristic while pruning nodes that violate the safety conditions induced by the static-map BRT $\mathcal{B}_\text{map}$ and human-robot BRT $\mathcal{B}_\text{rel}$.
In Line 11, the controller executes the selected control and updates the robot state, after which the same process is repeated until the robot reaches the goal.
The detailed mathematical definitions of three HJ value functions and pruning sets are presented in Sec.~\ref{sec:general-ttr}, Sec.~\ref{sec:map-brt} and Sec.~\ref{sec:rel-brt}.

To make explicit how the three precomputed HJ value functions guide online planning, Algorithm~\ref{alg:astar-example} presents an A*-based instantiation of the HJ-guided graph-search planner.
Let $x_\text{rob}(v)$ denote the robot state associated with a node $v \in \mathcal{V}$.
For a path $\{ v_0, v_1, ..., v_k\}$ from $v_0=v_{\mathrm{start}}$ to $v_k=v$, the cumulative transition cost is defined as $\bar{C}(v)=\sum_{j=0}^{k-1}\bar{c}(v_j,v_{j+1})$.
The node priority then follows the standard A* form $f(v)=\bar{C}(v)+h(v)$, where $h(v)=\Phi(\bar{x}_\text{rob}(v))$.
Here, for each expanded node $v$, the notation $v^{+}$ denotes a dynamics-feasible neighboring successor of $v$, namely a node satisfying $(v,v^{+}) \in \mathcal{E}_{\mathrm{feas}}$. 

\begin{algorithm}[htb]
    \caption{A*-Based Example of HJ-Guided Planning}
    \label{alg:astar-example}
    \begin{algorithmic}[1]
        \Require Directed graph $G=(\mathcal{V},\mathcal{E})$, start node $v_{\mathrm{start}}$, goal set $\mathcal{V}_{\mathrm{goal}}$, detected humans $\{x_\mathrm{hum}^i\}_{i\in\mathcal{I}}$, heuristic $h(v)=\Phi(\bar{x}_\text{rob}(v))$, pruning sets $\mathcal{B}_{\mathrm{map}}, \mathcal{B}_{\mathrm{rel}}$.
        \State Initialize OPEN with $v_{\mathrm{start}}$
        \State Set $\bar{C}(v_{\mathrm{start}})\gets 0$ and $f(v_{\mathrm{start}})\gets \bar{C}(v_{\mathrm{start}})+h(v_{\mathrm{start}})$
        \While{OPEN is not empty}
            \State Pop $v \in \mathrm{OPEN}$ with minimum key $f(v)$
            \If{$v \in \mathcal{V}_{\mathrm{goal}}$}
                \State \Return planned trajectory
            \EndIf
            \ForAll{$v^+$ such that $(v,v^+) \in \mathcal{E}_{\mathrm{feas}}$}
                \If{$x_\text{rob}(v^+) \in \mathcal{B}_{\mathrm{map}}$}
                    \State \textbf{continue}
                \EndIf
                \State $\mathrm{unsafe}\gets\mathrm{false}$
                \ForAll{$i \in \mathcal{I}$}
                    \State $x_{\mathrm{rel}}^i(v^+) \gets R(\theta(v^+))\big(x_{\mathrm{hum}}^i-Qx_\text{rob}(v^+)\big)$
                    \If{$x_{\mathrm{rel}}^i(v^+) \in \mathcal{B}_{\mathrm{rel}}$}
                        \State $\mathrm{unsafe}\gets\mathrm{true}$; \textbf{break}
                    \EndIf
                \EndFor
                \If{$\mathrm{unsafe}$}
                    \State \textbf{continue}
                \EndIf
                \State $\bar{C}(v^+) \gets \bar{C}(v)+\bar{c}(v,v^+)$
                \State $f(v^+) \gets \bar{C}(v^+)+h(v^+)$
                \State Insert or update $v^+$ in OPEN with key $f(v^+)$
            \EndFor
        \EndWhile
        \State \Return No feasible trajectory.
    \end{algorithmic}
\end{algorithm}

Algorithm~\ref{alg:astar-example} shows how the precomputed HJ value functions are used during online graph search.
The TTR value function defines the heuristic term in the A* priority, while the static-map and human-robot BRTs are used to prune unsafe successor nodes before insertion into the search frontier.

This HJ-Guided framework improves \textbf{search efficiency} in two complementary ways.
First, the TTR $\Phi(\bar{x}_\text{rob})$ serves as an informed heuristic that explicitly accounts for the robot dynamics, thereby guiding the search more effectively than purely geometric distance-based heuristics.
This dynamic-aware heuristic provides effective guidance during search, particularly in environments with kinematic constraints.
Second, the precomputed pruning conditions, $\mathcal{B}_\text{map}$ and $\mathcal{B}_\text{rel}$, remove predictively unsafe nodes before they are unnecessarily expanded, which reduces wasted search effort in both obstacle-cluttered and human-populated environments.
The framework also improves \textbf{safety awareness} in two complementary ways.
For static obstacles, the planner reasons not only about instantaneous collision but also about whether a state will inevitably evolve into collision under the robot dynamics with the static-map BRT $\mathcal{B}_\text{map}$.
For human interactions, the planner avoids relying on repeated online human-motion prediction and instead queries a precomputed relative-state human-robot BRT $\mathcal{B}_\text{rel}$, which preserves predictive safety reasoning while keeping each online node expansion lightweight.

\subsection{General TTR as the Guidance Heuristic}
\label{sec:general-ttr}

The first offline precomputed value function is a general TTR $\Phi : \bar{\mathcal{X}}_\text{rob}  \rightarrow \mathbb{R}_{\ge 0}$, computed in obstacle-free space by solving the HJI PDE Eq.~\eqref{HJI-TTR}.
Here, $\bar{x}_\text{rob}$ denotes the robot state expressed in a local coordinate frame with the goal placed at the origin.
At runtime, the current goal is handled by transforming the robot state $x_\text{rob}$ into this goal-centered frame $\bar{x}_\text{rob}$ before querying $\Phi$.
This construction yields a map-independent but dynamics-aware estimate of the minimum remaining time-to-reach.

\begin{figure}[htb]
    \centering
    \includegraphics[width=0.48\textwidth]{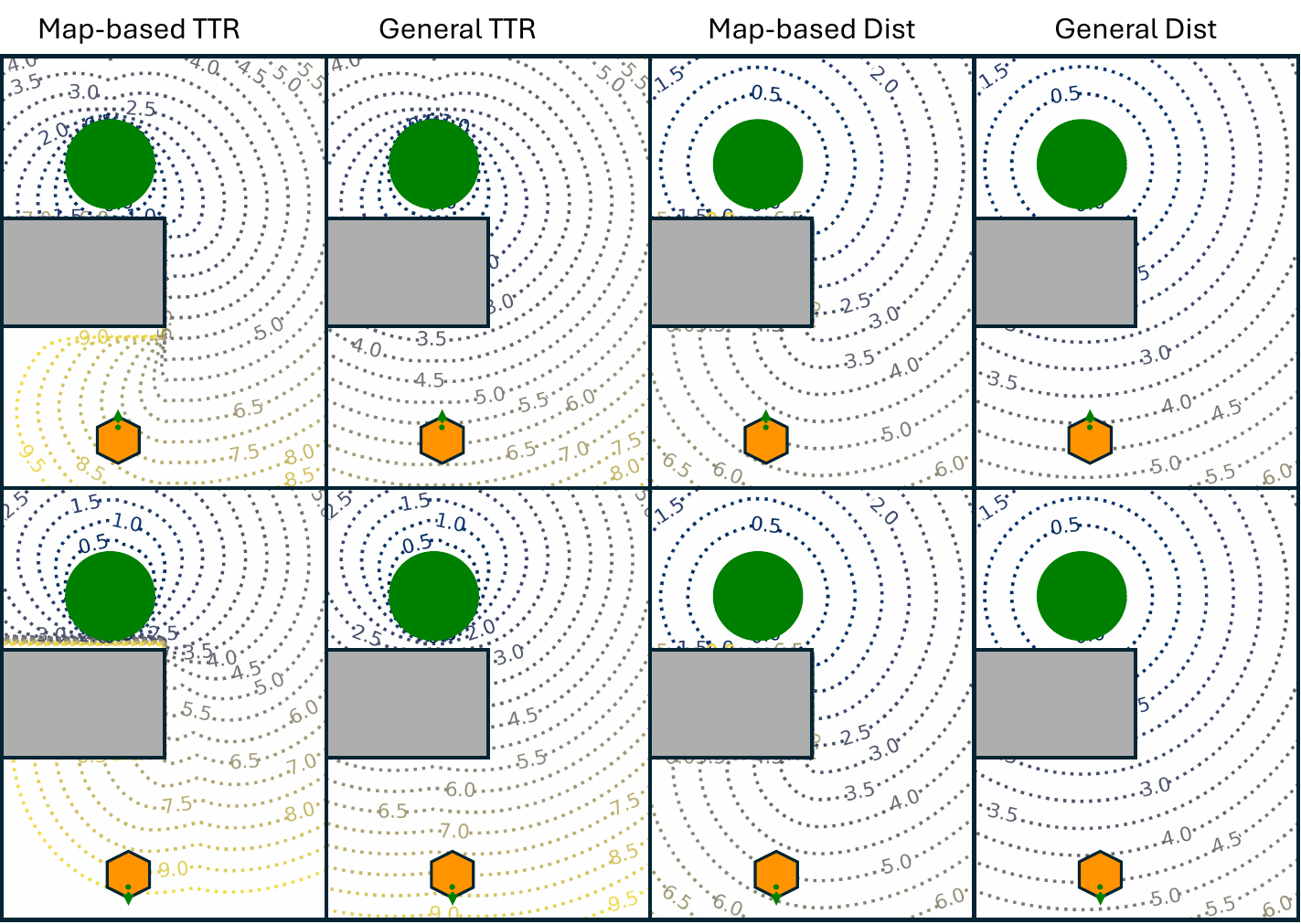}
    \caption{Heuristic comparisons for the robot at two orientations: map-based TTR, general TTR, map-based Dist, and general Dist heuristic. The hierarchy of values follows two rules: heuristics considering dynamics (TTR) are greater than those that do not (Dist), and map-based ones are greater than their general counterparts.}
    \label{heuristics}
\end{figure}

We consider two kinds of TTR heuristics, namely a map-based TTR and a general goal-centered TTR.
In theory, the map-based TTR is an exact heuristic for static environments, assuming perfect alignment between the real robot and its dynamical model. 
However, precomputing a map-based TTR is impractical when goal locations are not known prior to navigation.
As a result, we instead employ a general TTR $\Phi(\bar{x}_\text{rob})$ precomputed in obstacle-free space with the goal placed at the origin.
In online deployment, the robot state is transformed into the corresponding goal-centered frame before querying $\Phi$.
Unlike map-dependent value functions, this general TTR is environment-agnostic and can be reused across different environments and goal configurations without recomputation.
Although this map-free TTR does not encode static obstacle information and is therefore less informed than a map-specific value function, it remains highly effective as a guidance heuristic because it preserves the dynamical structure of the system.
For baseline comparison, we also include two distance-based heuristics (denoted as Dist): a map-based variant and a general variant.
The Dist heuristic is computed as the Euclidean distance to the goal divided by the robot's maximum speed.
The comparative contour plots of these heuristics are visualized for two robot states in Fig.~\ref{heuristics}.

The figure is organized into four columns, representing the map-based TTR, general TTR, map-based Dist, and general Dist, respectively.
The first row illustrates the heuristic contours for a robot located at $(2.0, 1.0)$ with a speed of $0.2\,\text{m/s}$ and an upward heading.
For this state, the heuristic values are 7.57 s, 5.81 s, 5.09 s, and 4.20 s separately. 
The second row shows the heuristics for the same position and speed, but with the robot facing in the opposite direction.
This reversed orientation substantially increases the TTR values, yielding 9.06 s for the map-based TTR and 7.89 s for the general TTR, while the distance-based heuristics remain unchanged at 5.09 s and 4.20 s, respectively.
Across both cases, the heuristics consistently satisfy the relation: the more informed heuristics yield greater values, with dynamics-aware (TTR) dominating distance-based (Dist) methods, and map-based versions dominating their general counterparts. 
This ordering is expected, as the map-based TTR is the most informed, because it provides the exact time estimation while respecting both dynamics and obstacles. 
The general TTR, which ignores static obstacles but takes the dynamics into consideration, provides an admissible but less informed underestimation of the exact map-based TTR.
Similarly, the map-based Dist heuristic, which considers obstacles but not dynamics, is more informed than the general Dist heuristic, which assumes an optimal straight line trajectory at maximum speed, disregarding both obstacles and dynamics.
Among these four heuristics, the general TTR achieves a favorable balance, retaining substantially greater informativeness than distance-based heuristics while remaining more practical than the map-based TTR.
Although the general TTR ignores static obstacles, potential misguidance is mitigated by the static-map BRT, which proactively prunes infeasible states during search.

\subsection{Static-Map BRT for Feasible Set Pruning}
\label{sec:map-brt}

The second offline precomputed value function is a static-map BRT defined on the robot state space $\mathcal{X}_\text{rob}$.
It is used to identify robot states that are predictively unsafe with respect to static obstacles and should therefore be pruned during search.
Given the static obstacle set $\mathcal{O}$ or equivalently the induced static map $\mathcal{M}$, we characterize the collision failure set $\mathcal{F}_\text{map}$ using an implicit surface function $l_{\mathcal{F}_\text{map}}(x_\text{rob})$, defined as the signed distance to the static obstacles inflated by the robot radius.
Thus, $l_{\mathcal{F}_\text{map}}(x_\text{rob}) \le 0$ indicates collision.

Assuming no external disturbances, we solve the HJI PDE in Eq.~\eqref{HJI-BRT} over the horizon $[-T_\text{map},0]$ to obtain the static-map BRT value function $V_\text{map}(x_\text{rob},t)$.
For a sufficiently large horizon $T_\text{map}$, the solution converges numerically to a stationary value function, and we therefore use the shorthand $V_\text{map}(x_\text{rob})$.
The static-map BRT $\mathcal{B}_\text{map}$ used for feasible-set pruning is defined as follows:
\begin{equation}
    \label{FS-Prune}
    \mathcal{B}_\text{map} = \{ x_\text{rob} | V_\text{map}(x_\text{rob}) < \epsilon_\text{map} \}
\end{equation}
where the threshold $\epsilon_\text{map}$ is typically set to zero or a small positive value to account for numerical approximation errors, model mismatch, and system delays.

\begin{figure}[htb]
    \centering
    \includegraphics[width=0.48\textwidth]{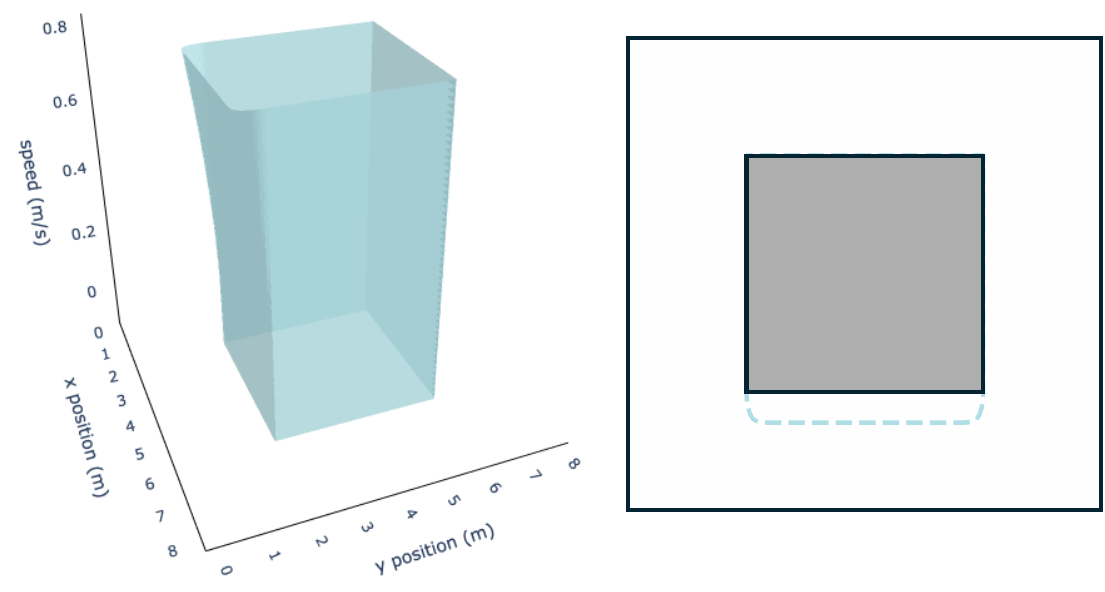}
    \caption{Visualization of one $\mathcal{B}_\text{map}$. (Left) The 3D value function for a fixed heading angle $\theta_{\text{rob}}= 1.57$ rad. (Right) A 2D slice of the $\mathcal{B}_\text{map}$ at $v_{\text{rob}}=0.75\, \text{m/s}$, $\theta_{\text{rob}}= 1.57$ rad corresponding to the plane indicated in the left plot.}
    \label{brts}
\end{figure}

This construction is consistent with the least-restrictive safety filter, because intervention is triggered only when the current state enters the BRT.
During online planning, any node that lies inside $\mathcal{B}_\text{map}$ is removed from the feasible state space before expansion.
Hence, the planner reasons not only about instantaneous collision with the map but also about whether a state is predictively unsafe under the robot dynamics within the prescribed horizon.
The geometry of $\mathcal{B}_\text{map}$ depends on both the robot state and the surrounding obstacles.
As illustrated in Fig.~\ref{brts}, the volume of the BRT increases with the robot's speed (positive $z$-axis direction), reflecting the reduced ability of the robot to avoid collisions at higher velocities.
This pruning strategy is particularly effective for static obstacles.
Since obstacle configurations remain unchanged over time, dangerous nodes can be identified and pruned as soon as their BRT values approach the safety threshold $\epsilon_{\text{map}}$, enabling proactive rather than reactive collision avoidance and maintaining safety well in advance of potential contact.

\subsection{Human-Robot BRT for Forward Invariant Pruning}
\label{sec:rel-brt}

The third offline value function is a human-robot BRT defined over the robot-centric relative state space $\mathcal{X}_\text{rel}$.
It is used to identify relative human-robot configurations that are predictively unsafe over a fixed horizon and should therefore be pruned during online planning.
To enable predictive safety reasoning in dynamic environments, we formulate a pursuit-evasion differential game between the robot and a human.
The relative state and dynamics are defined as:
\begin{equation}
    \label{Rel}
    \begin{aligned}
        x_\text{rel} &= R(\theta)(x_p - Qx_e) = R(\theta)(x_\text{hum} - Qx_\text{rob}), \\
        \dot{x}_\text{rel} &= g(x_\text{rel}, u_\text{rob}, u_\text{hum}),
    \end{aligned}
\end{equation}
where $R(\theta)$ is the rotation matrix aligning the human-robot relative position with the robot heading.
Here, $x_p$ and $x_e$ denote the pursuer ($x_\text{hum}$) and evader ($x_\text{rob}$) in the game respectively.
The operator $Q$ maps shared state components between $x_\text{hum}$ and $x_\text{rob}$ by either augmenting or reducing the state space of the robot.
The function $g$ denotes the resulting relative dynamics based on the relative state $x_\text{rel}$ under robot and human controls.
We characterize the human-robot collision failure set $\mathcal{F}_\text{rel}$ using an implicit surface function $l_{\mathcal{F}_\text{rel}}(x_\text{rel})$, defined as the signed separation distance between the robot and the human boundaries after accounting for their respective radii.
Thus, $l_{\mathcal{F}_\text{rel}}(x_\text{rel}) \le 0$ indicates collision.

By solving the HJI PDE in Eq.~\eqref{HJI-BRT} over the horizon $[-T_\text{rel},0]$, we obtain the value function $V_\text{rel}(x_\text{rel},t)$.
Since our online safety check is performed over a fixed prediction horizon, we use the shorthand $V_\text{rel}(x_\text{rel}) := V_\text{rel}(x_\text{rel},-T_\text{rel})$ for online pruning under a fixed horizon.

Rather than defining safety purely through a value threshold, we construct $\mathcal{B}_\text{rel}$ through a forward-invariant condition induced by the evolution of $V_\text{rel}(x_\text{rel})$.
This condition is closely related to Control-Barrier-Function-style (CBF) safety constraints applied to a precomputed reachability value function \cite{ames2019control,lu2025safe, borquez2024safety}.
Specifically, we define:
\begin{equation}
    \label{FI-Prune}
    \mathcal{B}_\text{rel} = \left\{ x_\text{rel}| \left. \frac{\partial{V}_\text{rel}}{\partial t} \right|_ {x_\text{rel}} < - \alpha ( V_\text{rel}(x_\text{rel}) - \epsilon_\text{rel}) \right\}
\end{equation}
where $\frac{\partial{V}_\text{rel}}{\partial t}$ is the partial derivative of the value function with respect to time $t$.
The scalar $\alpha$ is a positive scalar controlling the convergence rate of the value function.
The parameter $\epsilon_\text{rel}$ specifies a lower bound on acceptable safety margins.
This condition identifies states whose safety value is decreasing too rapidly and are therefore likely to violate the desired safety margin in the near future.

Since a graph search algorithm is employed, we can naturally use the finite difference between the neighbouring nodes and the current node to approximate the time derivative:
\begin{equation}
    \label{appro}
    \left. \frac{\partial{V}_\text{rel}}{\partial t} \right|_ {x_\text{rel}} \approx \frac{V_\text{rel}(x_\text{rel}^{\text{neb}}) - V_\text{rel}(x_\text{rel})}{\Delta t}
\end{equation}
where $\Delta t$ denotes the time interval between the current node and the neighbouring nodes $x_\text{rel}^{\text{neb}}$.
In our implementation, the time interval $\Delta t$ is a small value with a maximum time interval is 0.5 s, so the approximation Eq.\eqref{appro} holds.

The effectiveness of this Forward-Invariant (FI) pruning mechanism depends on the parameter $\alpha$, which governs the conservativeness of the safety constraint.
An overly small $\alpha$ will result in an excessively strict strategy, which may prune unnecessary nodes or leave none at all.
In converse, an overly large $\alpha$ reduces the method to a Feasible-Set (FS) pruning method, failing to preserve the predictive tendency of the value function. 
Moreover, while the pursuit-evasion game formulation provides a strong theoretical safety guarantee, its adversarial assumption can lead to overly conservative behavior in practice.
To balance theoretical safety guarantees with practical efficiency, we introduce an adaptive mechanism that dynamically adjusts $\alpha$ in response to the inferred level of collision risk.

The inferred collision risk is determined by two main factors: angular alignment ($a_{\text{align}}$) and distance alignment ($d_{\text{align}}$).
The angular alignment $a_{\text{align}}$ captures the tendency of the robot to orient itself toward the human.
It is formulated as an exponential similarity measure of the angular difference between the robot heading angle $\theta_{\text{rob}}$ and the line-of-sight vector from the robot to the human $\theta_{\text{r2h}}$:
\begin{equation}
    \label{a_align}
    a_{\text{align}} = \exp(- k \abs*{\theta_{\text{rob}} - \theta_{\text{r2h}}})
\end{equation}
where $k$ is a hyperparameter that scales the influence of the angular discrepancy.

The distance alignment $d_{\text{align}}$ models the closeness of the relative distance. 
It is formulated as a linearly decaying function of the relative distance $d_{\text{rel}}$ with respect to a predefined threshold $d_{\text{thre}}$:
\begin{equation}
    \label{d_align}
    d_{\text{align}} = \max \left( 0, 1 - \frac{d_{\text{rel}}}{d_{\text{thre}}}\right)
\end{equation}

\begin{figure}[htb]
    \centering
    \includegraphics[width=0.48\textwidth]{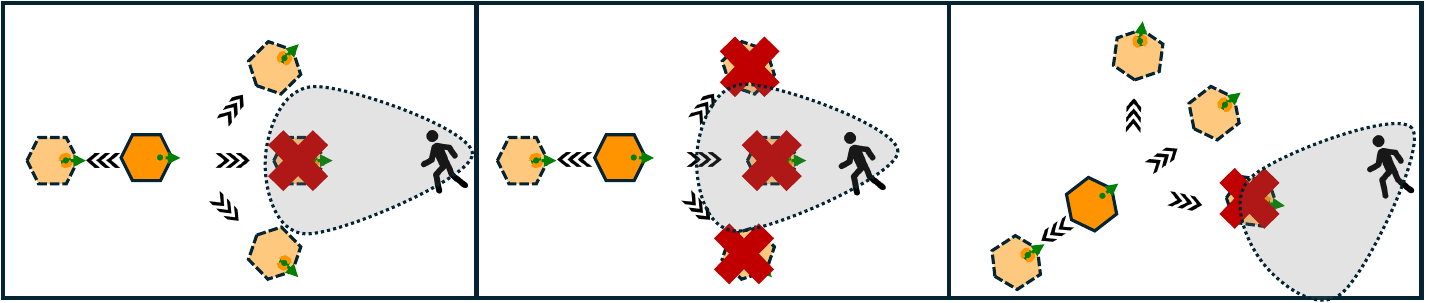}
    \caption{The influence of heading angle and relative distance on neighbor pruning. Compared to the baseline (left), a closer relative distance (middle) and a different heading angle (right) each result in a distinct set of pruned nodes, which are visualized with dashed outlines and reduced opacity.}
    \label{Rel-BRT}
\end{figure}

With Eq.~\eqref{a_align} and Eq.~\eqref{d_align}, the $\alpha$ is determined by:
\begin{equation}
    \alpha = \alpha_{\text{high}} - (\alpha_{\text{high}} - \alpha_{\text{low}}) (w_1 a_{\text{align}} + w_2 d_{\text{align}})
\end{equation}
where $w_1$ and $w_2$ are the linear weights of the two factors, 
$\alpha_{\text{low}}$ and $\alpha_{\text{high}}$ are determined by the discrete time step and neighboring nodes, with a theoretical range of $ \alpha \in [ \frac{0.1}{\Delta t},  \frac{1}{\Delta t}]$. 
As illustrated in Fig.~\ref{Rel-BRT}, different heading angles or relative distance will result in different pruning results.

We formally distinguish the proposed FI pruning method (Eq.~\eqref{FI-Prune}) from the conventional FS pruning strategy (defined as $\mathcal{B}_{\text{rel}} = \{x_{\text{rel}} \mid V_{\text{rel}}(x_{\text{rel}}) < 0\}$).
While the FS pruning relies on a binary check of whether a state currently resides within an unsafe set, it ignores the future evolution of safety.
This static nature can lead to delayed responses, as pruning is only triggered when the robot is already in close proximity to the danger boundary.
In contrast, the FI pruning evaluates the trend of safety by monitoring the time derivative of the value function. 
A state is declared unsafe when the value decreases faster than a prescribed exponential threshold, indicating inevitable future safety violation.
Hence, the FI pruning enables a smoother and more predictive pruning process.
Formally, the FS method can be interpreted as a limiting case of the FI method where $\alpha \to \infty$ (relaxing the derivative constraint).
Thus, FI constitutes a strictly stronger safety criterion that anticipates future constraint violations.
This enhanced predictive safety introduces a behavioral trade-off: the strict derivative condition may force the robot to take wider detours if space permits or wait longer in confined spaces compared to the FS, which allows for more aggressive maneuvers near obstacles.
Our adaptive $\alpha$ mechanism addresses this by bridging the gap between the predictive safety of the FI and the navigational flexibility of the FS based on real-time risk assessment.
The parameters of the FI-BRT, including $\alpha$, $\epsilon_{\text{rel}}$, and $\Delta t$, govern the conservativeness of the forward-invariant pruning strategy and are chosen based on the discretization resolution and the robot's dynamic limits.
The specific parameter values used for the FI-BRT are provided in Appendix~\ref{app:fibrt}.
Overall, the proposed FI-BRT framework enables the planner to eliminate states that are not only unsafe within the time horizon but provably destined to become unsafe, thereby introducing predictive safety reasoning into graph search.
\section{Simulation}
\label{sim}

The robot is modeled using a four-dimensional Dubins car dynamics.
Its state is defined as $x_{\text{rob}} = [x,\, y,\, v,\, \theta]^\top$, where $(x,y)$ denotes the planar position, $v$ the translational speed, and $\theta$ the heading angle.
The robot control is defined as $u_{\text{rob}} = [a,\, \omega]^\top$, where $a \in [-0.5,\,0.5]~\text{m/s}^2$ is the longitudinal acceleration and $\omega \in [-0.5,\,0.5]~\text{rad/s}$ is the turn rate.
Its dynamics are given by:
\begin{equation}
\label{eq:sim_robot_dynamics}
\begin{aligned}
\dot{x} &= v\cos\theta \\
\dot{y} &= v\sin\theta \\
\dot{v} &= a \\
\dot{\theta} &= \omega
\end{aligned}
\end{equation}

Each human is modeled as a planar point mass with constant speed $v_{\text{hum}}$.
The human state is given by its planar position $x_{\text{hum}} = [x_{\text{hum}},\, y_{\text{hum}}]^\top$, and its motion is parameterized by the direction angle $\theta_{\text{hum}}$.
In our simulations, we set $v_{\text{hum}} = 0.8~\text{m/s}$.
Accordingly, the human dynamics are
\begin{equation}
\label{eq:sim_human_dynamics}
\begin{aligned}
\dot{x}_{\text{hum}} &= v_{\text{hum}}\cos\theta_{\text{hum}}, \\
\dot{y}_{\text{hum}} &= v_{\text{hum}}\sin\theta_{\text{hum}}.
\end{aligned}
\end{equation}

We evaluated our proposed HJ reachability-guided framework on two graph search algorithms: A* \cite{hart1968formal} and ANA* \cite{van2011anytime}.
A* guarantees optimality when an admissible heuristic is used.
ANA*, an anytime variant of A*, aims to rapidly compute an initial feasible trajectory and subsequently refines the solution within the search time limitation.
Following the graph-search formulation in Section~\ref{sec:graph-hj-connection}, nodes are prioritized according to the sum of the cumulative transition costs along the current path and the heuristic estimate.
The cumulative cost-to-come for a trajectory \((v_0,\dots,v_k)\) is $\sum_{j=0}^{k-1} \bar{c}(v_j,v_{j+1})$.
In our implementation, the transition cost is instantiated using the static costmap \(M(\cdot)\) from Nav2 \cite{macenski2020marathon} as $\bar{c}(v_i,v_j)=M(v_j)$, where \(M(v_j)\in[0,100]\) denotes the static obstacle cost associated with node \(v_j\).

\begin{figure}[htp]
    \centering
    \includegraphics[width=0.48\textwidth]{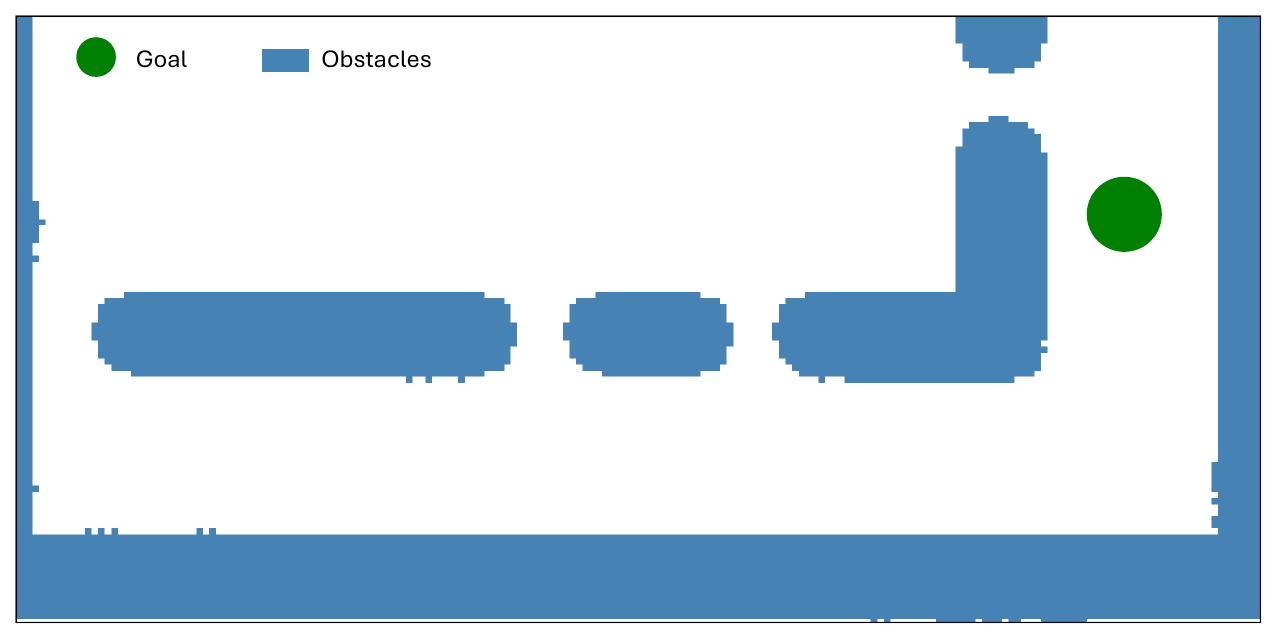}
    \caption{Simulated navigation costmap with static obstacles. Free space is shown in white; obstacles are shown in blue.}
    \label{costmap}
\end{figure}

The simulated environment is based on a 9.6 m $\times$ 5.4 m real-world room that contains multiple static obstacles as shown in Fig.~\ref{costmap}.
For each test, we specify a goal position as well as the initial state of the robot (position and heading angle). 
The robot is considered to have successfully arrived at the goal when the Euclidean distance between its current position and the goal is less than $0.2$ m for ANA*-based planners and $0.5$ m for A*-based planners.
Here we use a tighter goal tolerance for ANA* to reflect its anytime setting and shorter time budget, while a slightly looser tolerance is used for A* to ensure comparability under its larger computational burden.
To ensure safety during operation, we impose a speed limit of $0.1$ m/s upon reaching the goal position, preventing potential collisions in the vicinity of the goal.
We evaluate performance in terms of both path quality and search efficiency.
Path quality is measured by the total accumulated cost, computed only over successful trials to avoid distortions due to penalties from failed runs.
Search efficiency is quantified by the number of node expansions.
For A* and its variants, we enforce a $1 \times 10^6$ node expansion limit to bound computation time. 
The ANA* and its variants are evaluated slightly different: we record the number of node expansions required to first reach the goal within $0.04$ s search time limit, circumventing their incremental refinement nature. 
All tests are classified as either successes (timely goal achievement) or timeouts, ensuring objective performance comparison under realistic computational constraints.
Finally, to ensure path reliability, we conduct an additional feasibility check by interpolating the planned trajectories according to the system dynamics and verifying that the densified trajectories do not intersect with any obstacles.

First we evaluated the effectiveness of the \textbf{TTR heuristic} and the \textbf{BRT pruner} in two cases respectively.

\subsection{Heuristic Comparison}
\textbf{TTR heuristic} provides more accurate time estimation than the normal distance-based heuristic and thus it can improve the efficiency of searching process in most cases.
As a baseline comparison, we adopt a common heuristic (denoted as Dist): defined as the Euclidean distance to the goal divided by the maximum speed of the robot.
To show the effectiveness of the heuristic, we set the state cost to 1.0 for all the nodes in this test.

As for A*-based planners, we set the initial position at $(-2.6, 5.0)$ and the goal position at $(-2.6, 8.0)$. 
Initial heading angles are configured at $0^\circ$ , $90^\circ$ , $180^\circ$ , and $270^\circ$.
As shown in Table~\ref{tab: ttr-astar}, both planners finished all four tasks.
Notably, A* with TTR achieved this with significantly fewer node expansions ($6.483 \times 10^{5}$ vs. $1.291 \times 10^{7}$) and a much shorter total search time (20.97 s vs. 414.16 s) compared to A* with Dist.
In addition, we further measured the total evaluation time of each heuristic: TTR took 1.86 s (8.85\% of its total time), while Dist required 15.16 s (3.66\% of its total time).
This indicates that the performance improvement of the TTR heuristic stems primarily from reducing node expansions rather than accelerating heuristic evaluations.
Finally, both planners achieved identical average optimal costs (5.25), indicating that using the TTR heuristic did not compromise solution quality while substantially reducing the number of expansions.

\begin{table}[htp]
    \centering
    \caption{Comparison of Heuristics for A*.}
    \label{tab: ttr-astar}
    \begin{tabular}{ccccc}
    \toprule
    Method & Success & Nodes & Time & Costs \\ 
    \midrule
    A* w/ Dist           & 4/4 & $1.291 \times 10^{7}$         & 414.16           & 5.25 \\ 
    $\textbf{A* w/ TTR}$ & 4/4 & $\mathbf{6.483 \times 10^{5}}$ & $\textbf{20.97}$ & 5.25 \\ 
    \bottomrule
    \end{tabular}
\end{table}

To further contextualize these results, we assessed the feasibility of real-time deployment.
We benchmarked an industry-standard planner (Nav2 SmacPlanner \cite{macenski2020marathon, macenski2024open}) implemented in optimized C++ with a three-dimensional Dubins car dynamic.
Our tests indicate that such a system can process approximately $1.9 \times 10^{6}$ iterations per second.
Even with this high-performance execution speed, expanding the $1.291 \times 10^{7}$ nodes required by the Dist heuristic in our four-dimensional state space would theoretically consume over 6.8 seconds, which fails to meet the real-time requirements (typically $< 500$ ms) of the indoor navigation task.
In contrast, the node count reduced by our TTR heuristic ($6.483 \times 10^{5}$) would theoretically require only $\sim 0.34$ seconds in the same C++ environment.
This analysis suggests that the bottleneck lies in the algorithmic complexity (search space size) rather than implementation efficiency, highlighting the necessity of the TTR heuristic for real-time performance.

\begin{table}[htp]
    \centering
    \caption{Comparison of heuristics for ANA* on the common subset of 7 tasks successfully solved by both methods.}
    \label{tab: ttr-anastar}
    \begin{tabular}{cccc}
    \toprule
    Method & Success & Nodes$^\dagger$ & Costs$^\dagger$ \\ 
    \midrule
    ANA* w/ Dist           & 7/12             & \textbf{257.0}           & \textbf{8.06} \\ 
    \textbf{ANA* w/ TTR}   & \textbf{12}/12   & 258.6                    & 8.20 \\ 
    \bottomrule
    \multicolumn{4}{l}{\footnotesize $^\dagger$ Metrics averaged over the 7 tasks solved by both planners.}
    \end{tabular}
\end{table}

\begin{figure}[htp]
    \centering
    \includegraphics[width=0.48\textwidth]{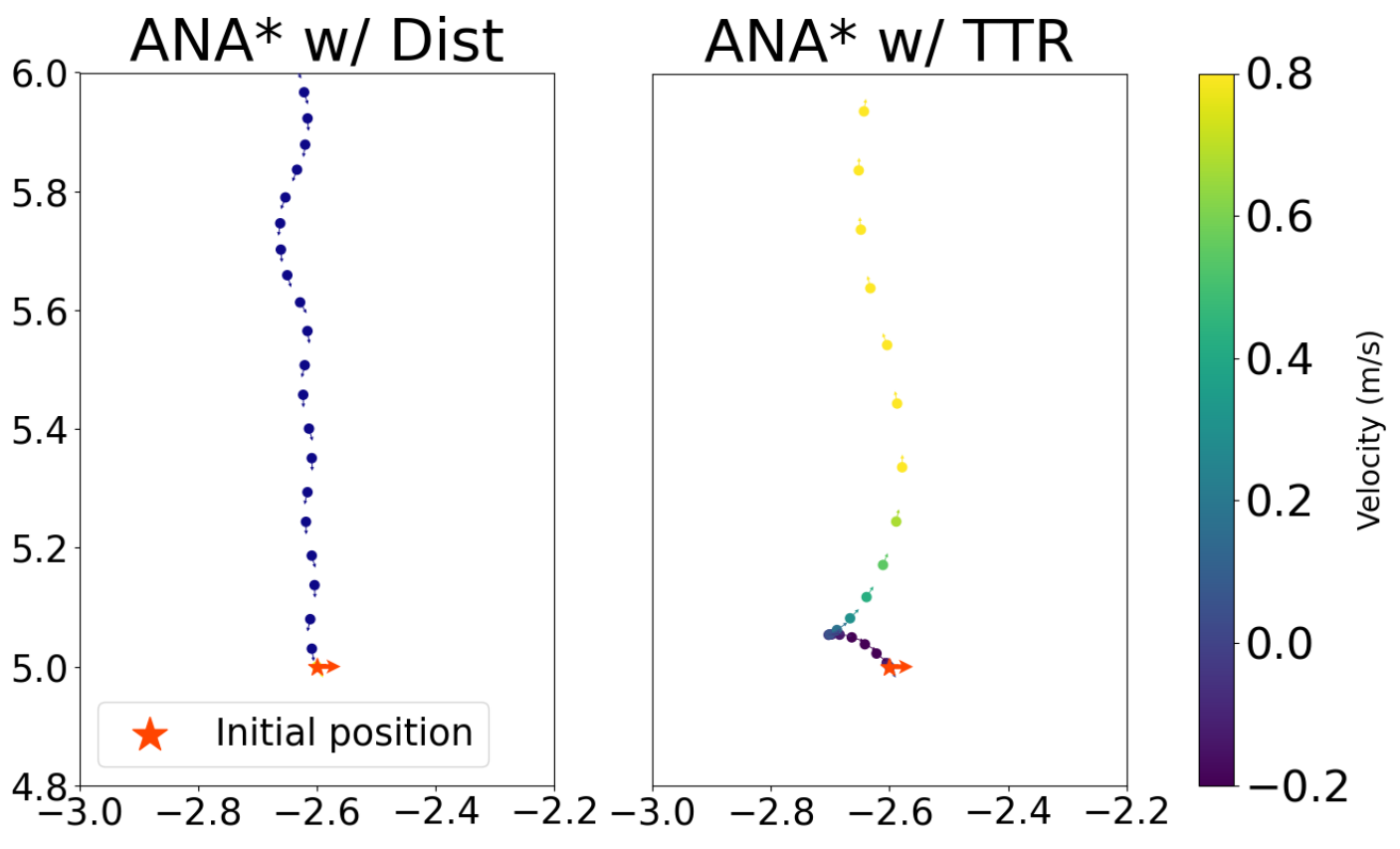}
    \caption{\textbf{Left}: ANA* w/ Dist created a trajectory requiring continuous backward movement (The heading angles represented by arrows are always pointing downwards). \textbf{Right}: ANA* w/ TTR generated a more reasonable trajectory where the robot reversed briefly and reoriented toward the goal.}
    \label{heuristic}
\end{figure}

As for ANA*-based planners, we set a longer-distance task where the initial position is at $(-2.6, 5.0)$ and the goal position at $(-2.6, 10.5)$.
To test the robustness against orientation, initial heading angles were set at $30^\circ$ intervals, creating a total of 12 distinct configurations.
As shown in Table~\ref{tab: ttr-anastar}, the choice of heuristic significantly impacted the success rate: while ANA* with the Dist heuristic failed in 5 out of 12 tasks, ANA* with the TTR heuristic successfully solved all 12 tasks.
The failures of the Dist heuristic occurred primarily when the robot was initially misaligned with the goal. 
Lacking dynamics awareness, the Dist heuristic guided the robot into local minima, resulting in persistent backward movements and eventual failure to reorient.
In contrast, ANA* with TTR demonstrated superior maneuverability: as illustrated in Fig.~\ref{heuristic}, it effectively guided the robot to execute a slight reversal followed by a precise reorientation toward the goal.
This confirms that the TTR heuristic provides critical guidance when non-holonomic constraints are significant.
To ensure a fair comparison, Table~\ref{tab: ttr-anastar} reports metrics averaged exclusively over the common subset of 7 tasks solved by both planners.
In these relatively simpler scenarios where orientation was less critical, both heuristics performed comparably: ANA* with Dist achieved almost similar node expansions (257.0 vs. 258.6) and marginally lower trajectory costs (8.06 vs. 8.20).
However, the advantage of the TTR heuristic becomes evident when considering the full dataset.
Across all 12 tasks—including the 5 complex cases where Dist failed, ANA* with TTR maintained a stable average node expansion of 273.4 and an average trajectory cost of 8.61.
This indicates that the TTR heuristic extends the planner's capability to solve complex orientation tasks with only a marginal increase in computational effort compared to the baseline.

\subsection{Pruning Strategies Comparison}
\textbf{BRT pruner} improves the search efficiency and trajectory safety by cutting off the nodes that will lead to collisions to the static obstacles.
We set the initial position at $(0.0, 4.8)$ and a closer goal position at $(0.2, 6.0)$ for A*-based planners due to the limited CPU memory issue.
As for ANA* planners, we set the initial position at $(-0.2, 4.8)$ the goal position at $(0.2, 7.2)$.
Both goal positions and initial positions are separated by the static obstacles and there is a traversable passage, allowing the robot a theoretically feasible path to cross the obstacle. 
We used the normal static obstacle pruner (denoted as the OBS pruner) as the baseline.
This kind of pruner will prune the nodes that lie in the static obstacles.
In addition, we used Dist heuristic to eliminate the potential TTR heuristic influences.
\begin{table}[htp]
    \centering
    \caption{Comparison of Pruning Strategies for A*.}
    \label{tab: prune-astar}
    \begin{tabular}{ccccc}
    \toprule
    Method & Success & Nodes & Time (s) & Costs \\ 
    \midrule
    A* w/ OBS            & 4/4 & $1.191 \times 10^{7}$         & $\textbf{303.32}$ & 89.5 \\ 
    $\textbf{A* w/ BRT}$ & 4/4 & $\mathbf{1.011 \times 10^{7}}$ & 335.84           & 89.5 \\ 
    \bottomrule
    \end{tabular}
\end{table}

As shown in Table~\ref{tab: prune-astar}, both A*-based planners finished all 4 tasks within the maximum iteration limit while maintaining the same optimal average costs of 89.5.
Specifically, A* with BRT pruner demonstrated superior search efficiency, reducing the average node expansions by 15.1\% ($1.011 \times 10^{7}$ vs. $1.191 \times 10^{7}$) compared to the OBS pruner baseline.
Despite this improvement in node expansion, the total searching time A* with BRT pruner is longer than A* with OBS pruner (335.84 vs. 303.32).
This is because BRT pruner operation itself required 41.97 s (12.5\% of its total time) which is substantially more than the 4.77 s (1.57\% of its total time) needed for the OBS pruner approach. 
This temporal analysis shows that, in our current implementation, the additional cost of evaluating the BRT pruner offsets the savings from reduced node expansions, resulting in a longer wall-clock planning time. 
Nevertheless, the reduction in node expansions is algorithmically meaningful and implementation-agnostic; it is therefore expected to translate into runtime gains in optimized implementations where reachability queries are accelerated (e.g., through efficient value-function lookup and batching).

\begin{table}[htp]
    \centering
    \caption{Comparison of pruning strategies for ANA* on the common subset of 8 tasks successfully solved by both methods.}
    \label{tab: prune-anastar}
    \begin{tabular}{cccc}
    \toprule
    Method & Success & Nodes$^\dagger$ & Costs$^\dagger$ \\ 
    \midrule
    ANA* w/ OBS            & 8/12              & 401.4                    & \textbf{404.57} \\ 
    \textbf{ANA* w/ BRT} & \textbf{12}/12  & \textbf{373.1} & 408.79 \\ 
    \bottomrule
    \multicolumn{4}{l}{\footnotesize $^\dagger$ Metrics averaged over the 8 tasks solved by both planners.}
    \end{tabular}
\end{table}

\begin{figure}[htp]
    \centering
    \includegraphics[width=0.48\textwidth]{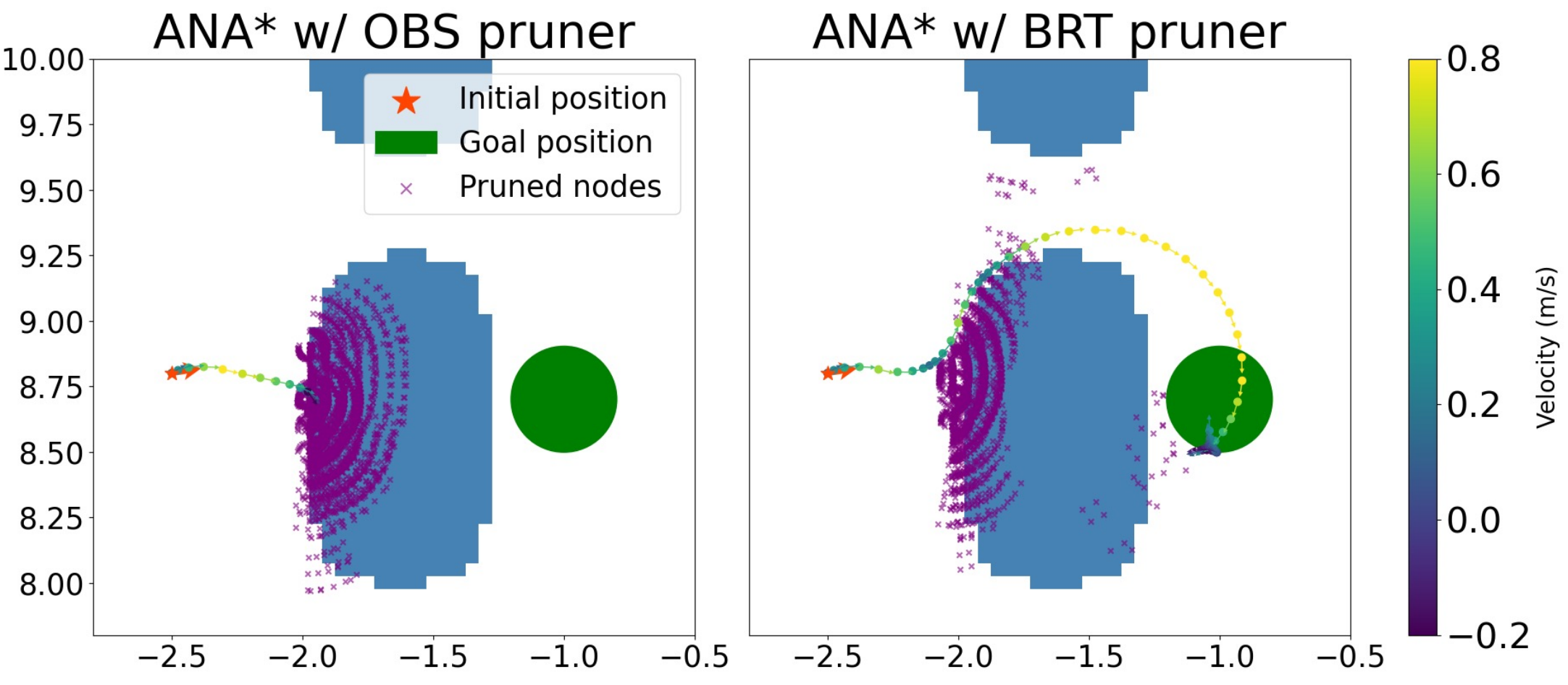}
    \caption{\textbf{Left}: ANA* w/ OBS pruner got too close to the obstacle to find the goal; \textbf{Right}: ANA* w/ BRT pruner eliminated dangerous nodes early and found the goal.}
    \label{pruner}
\end{figure}

As for ANA*-based planners, we compared the effects of two pruning strategies under the same long-distance setting with 12 initial groups.
As shown in Table~\ref{tab: prune-anastar}, ANA* equipped with the OBS pruner failed in 4 out of the 12 tasks, whereas the version with the BRT pruner successfully solved all 12.
The OBS pruner failed the feasibility interpolation check because densified trajectories intersected static obstacles.
To illustrate this issue, we show a task where the robot starts at $(-2.5, 8.8)$ with the heading angle 0.52 rad and aims to reach the goal at $(-1.0, 8.7)$ in Fig.~\ref{pruner}.
In this case, the planner with the OBS pruner approached the obstacle too closely and failed to circumvent it within the time limit. 
By contrast, the BRT-based planner avoided the obstacle earlier and successfully found a feasible path.
To ensure a fair quantitative comparison, Table~\ref{tab: prune-anastar} reports metrics averaged exclusively over the subset of 8 tasks successfully solved by both planners.
Within this common subset, ANA* with the BRT pruner expanded fewer nodes on average (373.1 vs. 401.4) while achieving similar path quality (408.79 vs. 404.57).
Notably, the efficiency of the BRT pruner remains robust even when considering the full dataset.
Across all 12 tasks (including the 4 tasks where the OBS failed), the BRT pruner maintained a low average node expansion of 326.8 and an average trajectory cost of 371.43.
These results indicate that the BRT pruner not only guarantees higher success rates by enabling safer maneuvers but also maintains high search efficiency across both simple and complex scenarios.

\subsection{General Static Test}
We further conducted a general evaluation by randomly sampling 100 initial-goal pairs, where each pair was either separated by obstacles or at least $1.0$ meters apart to ensure non-trivial navigation scenarios.
In this experiment, we focus on ANA*-based planners, as their anytime property enables efficient large-scale evaluation under strict time constraints, while A*-based planners were excluded due to their significantly higher computational cost in this setting.

\begin{table}[htp]
    \centering
    \caption{Performance of ANA* with different heuristic and pruner combinations on the common subset of 70 tasks successfully solved by all configurations.}
    \label{tab: random-ablations}
    \begin{tabular}{ccccc}
    \toprule
    Heuristic & Pruner & Success & Nodes$^\dagger$ & Costs$^\dagger$ \\ 
    \midrule
      Dist    &  OBS & 81/100 & 311.1 &  191.44 \\ 
      Dist    &  \textbf{BRT} & 85/100 & 299.9 &  188.98 \\ 
      \textbf{TTR}  &  OBS   & 83/100 & 235.4 &  146.58 \\ 
      \textbf{TTR}  & \textbf{BRT} & \textbf{97}/100 & \textbf{229.3} & \textbf{146.37} \\ 
    \bottomrule
    \multicolumn{5}{l}{\footnotesize $^\dagger$ Metrics averaged over the common subset of 70 tasks.}
    \end{tabular}
\end{table}

The results comparing four ANA*-based planners are summarized in Table~\ref{tab: random-ablations}.
Evidently, the proposed combination of the TTR heuristic and the BRT pruner achieves the superior overall performance, successfully completing \textbf{97 out of 100 tasks}—the highest success rate among all tested settings.
To ensure a rigorous assessment of search efficiency and path quality, Table~\ref{tab: random-ablations} reports metrics averaged exclusively over the common subset of 70 tasks that were successfully solved by all four combinations.
A detailed analysis of this subset reveals the distinct, complementary roles of the heuristic and the pruner.
The \textbf{TTR heuristic}, benefiting from its accurate estimation based on the dynamics, emerges as the dominant factor for improving search efficiency and path quality. 
For instance, when paired with the BRT pruner, replacing the Dist heuristic with TTR reduces the average node expansions by 23.5\% (from 299.9 to 229.3) and lowers the trajectory cost from 188.98 to 146.37.
A similar strong trend is observed with the OBS pruner: TTR improves the success rate from 81/100 to 83/100, decreases node expansions from 311.1 to 235.4, and reduces trajectory cost from 191.44 to 146.58.
The \textbf{BRT pruner} primarily enhances robustness and success rate, particularly in complex scenarios navigating around obstacles is critical, though its contribution to path quality is more nuanced compared to the TTR heuristic due to its computational overhead.
While its impact on trajectory cost is modest (e.g., maintaining comparable costs of 146.37 vs. 146.58 under the TTR heuristic), its contribution to the success rate is substantial, particularly in complex environments.
Most notably, when using the TTR heuristic, upgrading from the OBS pruner to the BRT pruner boosts the success rate from 83\% to 97\%.
A similar result is observed with the Dist heuristic: the BRT pruner improves the success rate from 81\% to 85\%.
It is worth noting that even when considering the full set of 97 tasks solved by the proposed method (including 27 complex scenarios failed by others), the average node expansion remains efficient at 287.2 at an average cost of 180.47.
In summary, these findings highlight a synergistic effect: the TTR heuristic provides a fundamental improvement in search guidance, while the BRT pruner ensures safety and feasibility. 
Their integration offers the most balanced trade-off, delivering high efficiency without compromising robustness.

\subsection{Dynamic Test}
\label{sim-dynamic}
The human-robot BRT pruner, implemented using the forward-invariant human-robot BRT $\mathcal{B}_{\text{rel}}$ in Eq.~\eqref{FI-Prune} (denoted as the \textbf{FI-BRT}), prunes nodes that are deemed unsafe.
To demonstrate the enhancement of the safety awareness, we design two simulation tests.
In these tests, the human moves with a constant 0.8 m/s speed and the human does not do any reaction to the robot.
So the robot needs to generate safe trajectories to avoid the collision.
We compare the proposed FI-BRT with:
\begin{itemize}
    \item \textbf{OBS Pruner}: Treats the human as a static obstacle at its instantaneous position during each planning iteration, representing a non-reachability-based baseline.
    \item \textbf{FS-BRT} (ablation): A feasible-set pruning strategy that uses the same human-robot BRT value function $V_{\text{rel}}$ but applies a static level-set condition without enforcing forward invariance.
\end{itemize}

In the first Face-to-Face scenario, the robot and a single human  are initially positioned along a straight line, facing each other. 
Their paths are configured to overlap, with the robot's start and end points encompassing those of the human. 
Both agents begin moving simultaneously. 
Under this configuration, the robot is required to adopt a collision avoidance strategy to bypass the human.
Specifically, the robot starts at $(0.0, 12.5)$ and its goal is $(0.0, 5.2)$; while the human starts at $(0.0, 6.0)$ and its goal is $(0.0, 12.4)$.

In the second Zig-Zag scenario, each human starts from the position $(0.0, 6.0)$ and moves at a constant velocity of 0.8 m/s toward the destination at $(0.0, 12.4)$. 
Upon reaching the destination, a new human is immediately spawned at the $(0.0, 6.0)$, resulting in a continuous stream of humans along the same trajectory.
The robot is tasked with repeatedly navigating through a cycle of four goals: $(-0.8, 12.0)$, $(0.8, 10.5)$, $(-0.8, 9.0)$, and $(0.8, 7.5)$. 
These goals are symmetrically distributed on either side of the human path, forming the robot's patrol route.
The robot visits the goals sequentially and, upon reaching the final goal, returns to the first to repeat the cycle.

\begin{table}[h]
    \centering
    \caption{Comparison of Pruning Strategies for Human}
    \label{tab: human-result}
    \begin{tabular}{ccccc}
    \toprule
    \multirow{2}{*}{\textbf{Methods}} & \multicolumn{2}{c}{Face-to-Face} & \multicolumn{2}{c}{Zig-Zag} \\ 
    \cmidrule(lr){2-3} \cmidrule(lr){4-5}
                      & Success & Avg. Time (s) & Success & Avg. Time (s) \\ 
    \midrule
    \textbf{FI-BRT}            & \textbf{5}/5 & \textbf{15.64} & \textbf{5}/5 & 39.85 \\ 
    \textbf{FS-BRT}            & 0/5          & Failed           & \textbf{5}/5 & 31.17 \\ 
    OBS               & 0/5          & Failed           & 1/5          & \textbf{30.41} \\ 
    \bottomrule
    \end{tabular}
\end{table}

The performance comparison of three pruning strategies across the two test scenarios is summarized in Table~\ref{tab: human-result}.
The results highlight a fundamental distinction between predictive safety, as enabled by FI-BRT, and reactive safety, as exhibited by FS-BRT.

\begin{figure}[htb]
    \centering
    \includegraphics[width=0.48\textwidth]{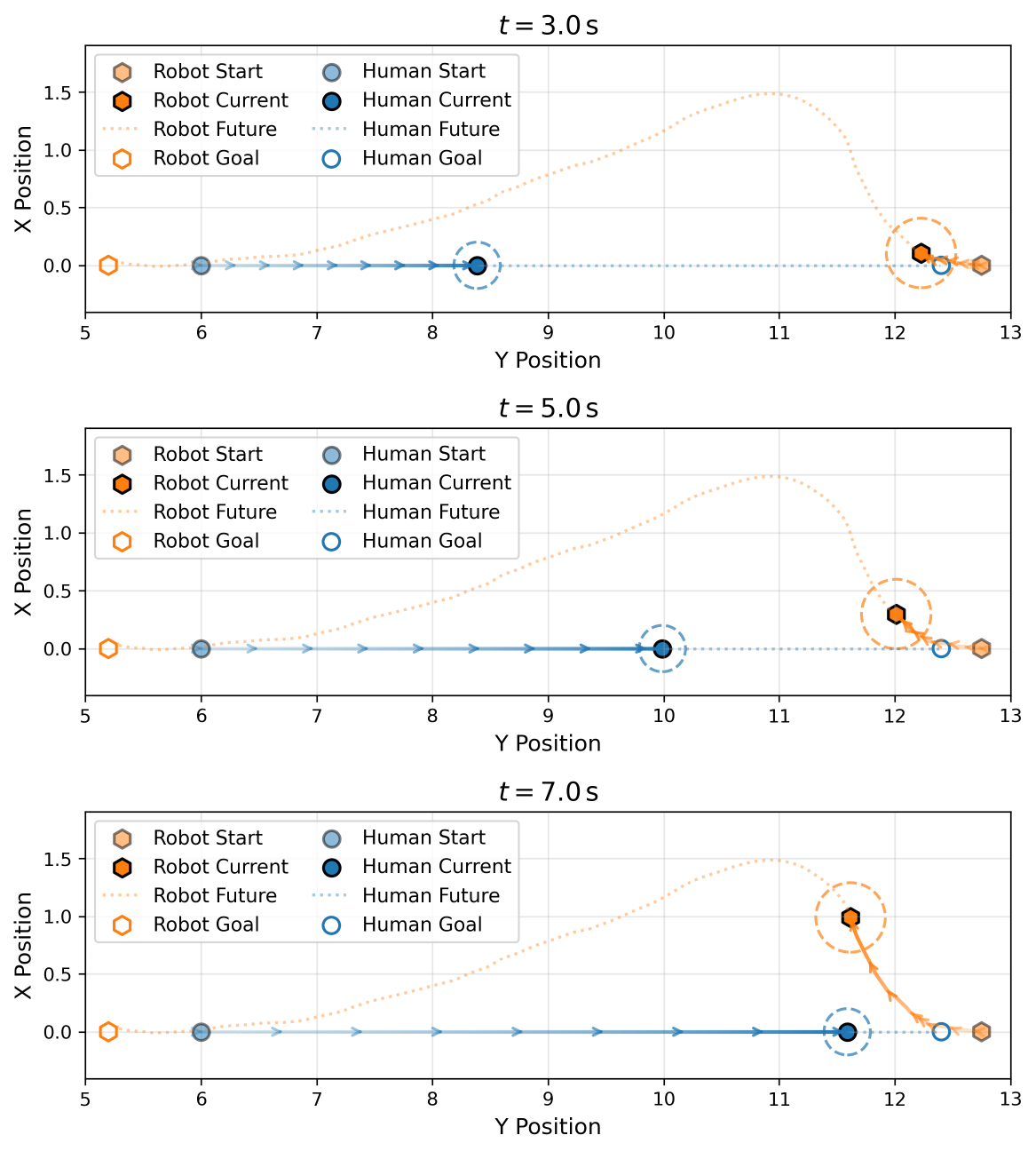}
    \caption{A sequence from the Face-to-Face scenario with the FI-BRT method demonstrates the robot's successful behavior to circumvent an oncoming human, with snapshots taken at t=3.0, 5.0, and 7.0 seconds.}
    \label{Human-F2F}
\end{figure}

In the Face-to-Face scenario, the FI-BRT completed all 5 tests. 
As illustrated in Fig.\ref{Human-F2F}, the robot starts to detour when the relative distance is still approximately 3.5 m. 
This early response is triggered because the FI-BRT identifies an unsafe rate of change of the value function, even though the current state may still lie outside the unsafe set.
In contrast, both the OBS baseline and the FS-BRT ablation failed completely (0/5). 
The failure of FS-BRT is particularly instructive: since it only prunes nodes strictly inside the unsafe set, it delays the evasion maneuver until the robot is close to the human.
Given the robot's kinematic constraints, this reactive response is often too late to avoid a collision in a head-on encounter.

In the Zig-Zag scenario, the trade-off between safety and efficiency becomes more evident.
The OBS pruner recorded the shortest average time (30.41 s) simply because it performs the least aggressive pruning, filtering out only immediate collision states without considering future dynamics.
However, this myopia comes at the cost of safety, resulting in a high failure rate (only 1/5 success).
Both FS-BRT and FI-BRT take longer navigation times compared to OBS, as they both incorporate a prediction horizon into the safety check.
Comparing the two, FS-BRT (31.17 s) is faster than FI-BRT (39.85 s).
The reason lies in the nature of their constraints: FS-BRT employs a binary check based on the value function's level set. 
This allows the robot to "skim" the boundary of the unsafe set, traversing tight spaces as long as the state remains theoretically feasible within the horizon.
In contrast, FI-BRT imposes a stricter condition on the rate of change of the value function.
Even if a state is currently outside the unsafe set (safe in FS terms), FI-BRT may still prune it if the trajectory leads towards danger too rapidly.
This effectively enforces a "safety buffer" in the time domain, compelling the robot to take wider detours or wait longer to ensure the safety trend remains positive, leading to the increased total time.

These results highlight two key insights. 
First, predictive safety enabled by FI-BRT is essential for avoiding head-on collisions, where reactive methods often fail. 
Second, enforcing forward invariance yields improved robustness at the cost of increased travel time, revealing an explicit safety--efficiency trade-off.

\subsection{Computational Time Efficiency Analysis}
The previous sections have evaluated the effectiveness of the different heuristics and pruning strategies, measuring their performances on node expansions and trajectory costs.
This section shifts the focus to their computational time efficiency.
To isolate and quantify the computational time, we conduct a dedicated study outside of the navigation simulation task.
We used the same dimensions of input variables and the same HJ value functions to to ensure fair and reliable comparisons.
Generally speaking, the search algorithm will evaluate the heuristics of all neighboring nodes and check the feasibility of all neighboring nodes (and the trajectory segments leading to these nodes) once every time it expands the current node.
Therefore, the real number of the function executions required for each task is determined by the search difficulty of the task and whether there is a search time limit.
To obtain measurable and stable time intervals, each function was executed 10,000 times in a tight loop in our analysis.
\begin{table}[htp]
    \centering
    \caption{Computational duration of heuristics and pruning modules. The reported BRT pruner timing applies to both static-map and human-robot BRTs.}
    \label{tab: time}
    \begin{tabular}{cccc}
    \toprule 
    Component & Input (num, dim) & Total Duration (s) & Rel. Speed \\ 
    \midrule
    TTR Heuristic  & (35, 4)  & 0.187 & 1.0x \\ 
    Dist Heuristic & (35, 2)  & 0.053 & 3.5x \\ 
    BRT Pruner     & (140, 4) & 0.247 & 1.0x \\ 
    OBS Pruner     & (140, 2) & 0.076 & 3.3x \\ 
    \bottomrule
    \end{tabular}
\end{table}

Table~\ref{tab: time} reports the computational time of different heuristics and pruners. 
The TTR heuristic, which processes inputs with 35 neighboring nodes and four dimensions per node, runs 3.5 times slower than the two-dimensional Dist heuristic (35 nodes, two dimensions per node). 
This performance difference is primarily because TTR requires converting input states from world to map coordinates, in addition to handling the two extra dimensions per node. 
Similarly, the BRT pruner, which processes 140 data points—comprising neighboring nodes and the trajectory segments leading to those nodes, each with 4 dimensions—exhibits a 3.3x slowdown compared to the OBS pruner (the same 140 data points, but 2 dimensions per point) for analogous reasons.
The same computational profile applies to the human-robot BRT pruner, which shares identical input dimensionality and evaluation structure with the static-map BRT pruner.
Overall, four-dimensional TTR and BRT pruner provides more accurate or safe evaluation results, but also are notably more expensive than their two-dimensional counterparts.
This provides another view to interpret the trade-offs between performance quality and computational expense observed in the prior results.
\section{Experiments}

To validate real-world applicability, we evaluate the performance of the proposed HJ reachability-guided search framework with the ANA* algorithm through a series of real-world experiments.
In addition, we also conducted ablative studies comparing our method against baseline methods.
Detailed video demonstrations of the experiments, including both static and dynamic scenarios, are provided in the supplementary material.

\subsection{Experimental Setup}
\textit{Test Environment}: 
We evaluated our method in both static and dynamic navigation scenarios within a 9.6 m $\times$ 5.4 m room.
Static tasks involved navigation in the presence of stationary obstacles only, whereas dynamic tasks introduced humans moving around within the workspace.
The position of the robot is estimated with an adaptive Monte-Carlo localization method by the Nav2 package \cite{macenski2020marathon},
and the human positions are estimated via 3D object detection features from the ZED 2 camera fixed on the top of the wall at approximately 30 Hz.
For each room configuration, a static costmap is constructed using Cartographer \cite{hess2016real}, an open-source SLAM framework developed by Google, which processes data from the onboard 2D LiDAR.

\textit{Hardware Platform:}
We conducted real-world experiments using the Quanser QBot platform \cite{QuanserQBotPlatform}, a differentially driven ground robot featuring a hexagonal base with a diameter of 570 mm and a height of 227 mm.
The robot is equipped with a 360-degree 2D LiDAR, an Intel RealSense D435 RGB-D camera, and onboard inertial sensors. 
To ensure consistency with the simulation settings described in Section~\ref{sim}, the robot's acceleration, speed, and turn-rate limits were conservatively chosen to match those assumed in simulation.
These limits reflect the physical capabilities of the platform while maintaining compatibility with the planning model.

\textit{Control Architecture:}
We adopt a segregated control architecture in which both the online planning module and the low-level control module are executed onboard the robot on two separate NVIDIA Jetson Orin Nano computers.
In our experiments, the online planner operates in a receding-horizon manner at a frequency of 12\,Hz. 
At each iteration, the planner solves the search problem and transmits the resulting action sequence to the controller via ROS~2 communication. The robot then executes its first several commands at a higher control frequency of 25\,Hz without modification.

\textit{Performance Metrics:}
Due to the iterative nature of the receding horizon strategy, monitoring intermediate planning metrics, such as the number of expanded nodes or the cost of individual search results, is not a meaningful measure of overall performance.  
Accordingly, we evaluate real-world performance using task-level outcomes, specifically the final navigation success and the total navigation time required to complete a fixed number of patrol tasks.
Specifically, the robot is required to patrol among several goals continuously while avoiding both static and dynamic obstacles. 
A single traversal from one goal to the next is considered successful only if it is completed without collision, deadlock, or external intervention.
For each tested method, we conducted three full patrol cycles (start and return the same goal is considered as a full patrol cycle) and recorded the number of successful traversals and the total time required to complete the patrol.

\textit{Offline HJ Value Functions Precomputation:}
All HJ value functions were precomputed offline on a desktop workstation equipped with an AMD Ryzen 9 3950X 16-core CPU.
The TTR value function occupies around 1.4 GB of memory and requires about 27.7 minutes to compute.
The static-map BRT has a size of 1.2 GB and is generated in about 26 minutes.
The human-robot BRT is significantly more compact, occupying only 23.9 MB, and is computed in about 5 min.

\subsection{Navigation in Static Environments}
\begin{figure}[htp]
  \centering
  \includegraphics[width=0.48\textwidth]{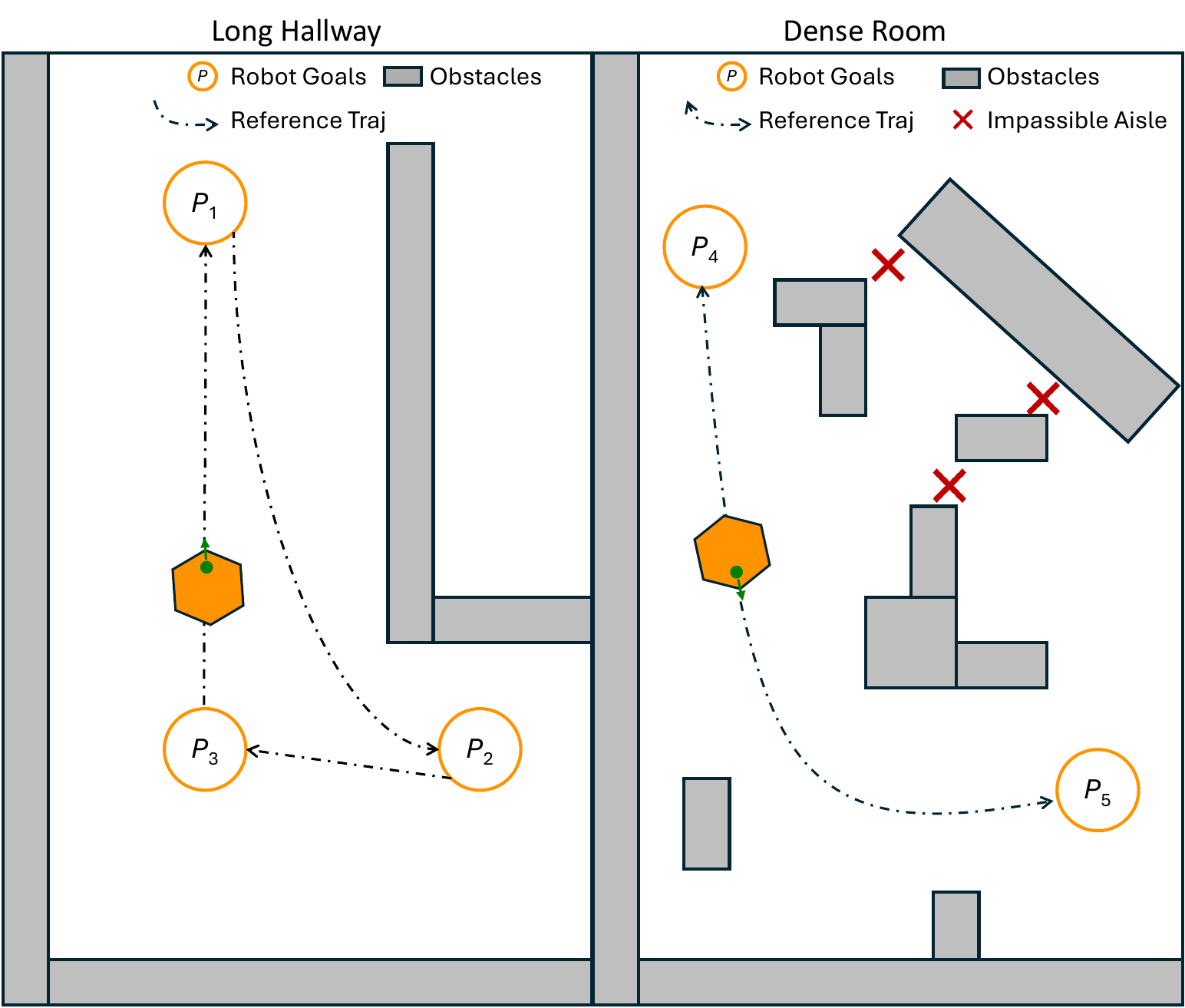}
  \caption{Layout of the two static environments: Long Hallway (left) and Dense Room (right). Dotted lines represent the reference trajectories connecting the goals ($P_1$-$P_5$). Note that the aisles marked with red crosses in the Dense Room are physically impassible for the robot.}
  \label{static-maps}
\end{figure}

Our initial experiments evaluate robotic navigation in static environments under two distinct obstacle configurations, referred to as the Long Hallway and the Dense Room as illustrated in Fig.~\ref{static-maps}.  
Quantitative results are summarized in Table~\ref{tab:exp1}.
In the Long Hallway configuration, three goal locations are specified: $P_1$ (1.2, 5.0), $P_2$ (2.7, -0.7), and $P_3$ (1.2, -0.7). 
The navigation sequence is cyclic: $P_1 \to P_2 \to P_3 \to P_1$.
In the Dense Room configuration, two goals are placed at $P_4$ (1.0, 4.2) and $P_5$ (3.0, -1.0), a total 5.6 m apart, to intentionally eliminate any direct line of sight, necessitating complex circumnavigation of static obstacles.

\begin{figure*}[htbp]
  \centering
  \includegraphics[width=1.0\textwidth]{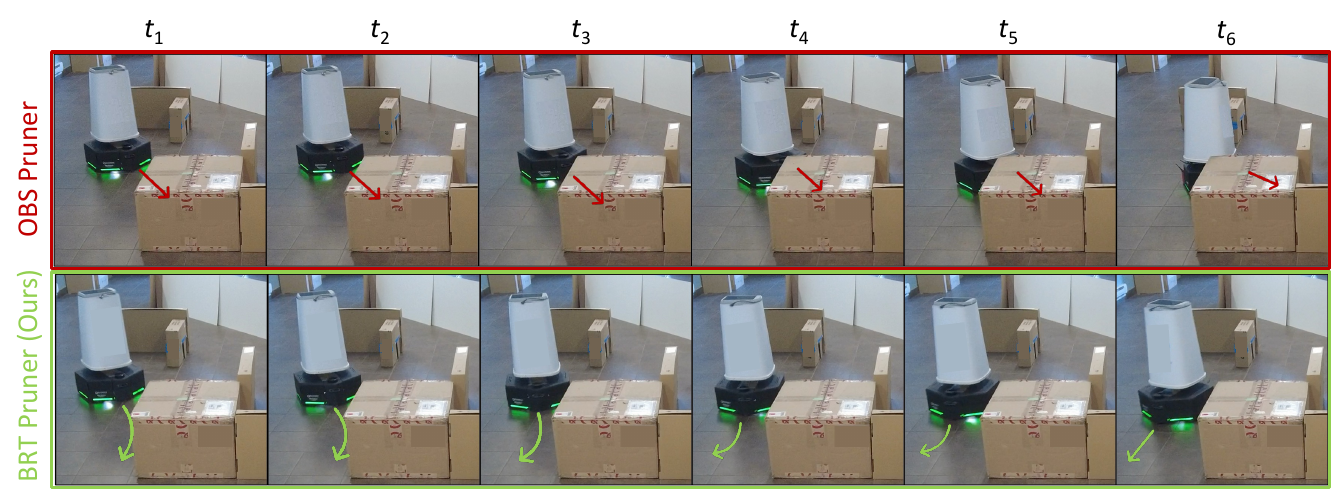}
  \caption{Comparison of the OBS and BRT pruners across consecutive timesteps (from $t_1$ to $t_6$) in the Dense Room scenario. \textbf{Top row (OBS Pruner):} Highlighted in the red box, the robot failed to prune dangerous nodes in time, causing it to plan too close to the obstacle and stall without finding a feasible trajectory within the time limit, as indicated by the red arrows (robot heading). \textbf{Bottom row (Ours):} Highlighted in the green box, our BRT pruner effectively identifies and eliminates dangerous nodes early, allowing the robot to successfully circumvent the obstacle shown by the green arrows.}
\label{exp:compa1}
\end{figure*}

In the Long Hallway scenario, the performance gap is primarily dictated by the choice of heuristic rather than the pruner. 
As shown in Table~\ref{tab:exp1}, planners employing the Dist heuristic struggled significantly, achieving a low success rate of only 2/9. 
The inclusion of our BRT pruner yielded no improvement in success rate over the OBS pruner for the Dist heuristic, although it slightly reduced the navigation time, suggesting that simple Euclidean distance guidance is insufficient for this specific arrangement. 
Conversely, the TTR heuristic proved highly effective, achieving a perfect 9/9 success rate with both pruners. 
In addition, while both TTR-based methods succeeded, the combination with the BRT pruner demonstrated improved efficiency, reducing the total navigation time to 85.9 s compared to 89.6 s with the OBS pruner. 
This indicates that the TTR heuristic can provide necessary global guidance for long-range navigation, and the BRT pruner can contribute to efficiency improvement.

\begin{table}[htbp]
  \centering
  \caption{Navigation performance in static environments. For methods with partial success, the reported time corresponds to successful trips only.}
  \label{tab:exp1}
  \begin{tabular}{cccccc}
    \toprule
    \multirow{2}{*}{Heuristic} & \multirow{2}{*}{Pruner} & \multicolumn{2}{c}{Long Hallway} & \multicolumn{2}{c}{Dense Room} \\
    \cmidrule(lr){3-4} \cmidrule(lr){5-6}
    & & Success & Time (s) & Success & Time (s) \\ 
    \midrule
    Dist          & OBS          & 2/9          & 29.8 & 0/6          & Failed   \\ 
    Dist          & \textbf{BRT} & 2/9          & 24.5 & 1/6          & 20.1   \\ 
    \textbf{TTR}  & OBS          & \textbf{9}/9 & 89.6 & 0/6          & Failed  \\
    \textbf{TTR}  & \textbf{BRT} & \textbf{9}/9 & 85.9 & \textbf{6}/6 & 115.6 \\ 
    \bottomrule
  \end{tabular}
\end{table}

As listed in Table~\ref{tab:exp1}, the Dense Room scenario presented a greater challenge.
All planners utilizing the basic OBS pruner failed completely, achieving a 0/6 success rate irrespective of the heuristic in the Dense Room scenario.
This resulted in a computation time reported as Failed, indicating no trip was completed out of the total 6 trips.
The planner using the Dist heuristic and the BRT pruner achieved a marginal success rate of 1/6.
Its failure during the return journey highlights a key limitation of the Dist heuristic: its inability to provide effective guidance in complex geometric scenarios, ultimately leading it to become trapped at a critical corner.
The single successful one-way trip required 20.1 s of navigation time.
In stark contrast, the synergistic combination of the TTR heuristic and the BRT pruner (our method) demonstrated superior efficacy. 
This method is the only one to achieve a 100 $\%$ success rate (6/6) completing all trial trips with a total navigation time of 115.6\,s.
To elucidate the reason for this performance gap, we conducted a detailed analysis of the planning process for both success and failure cases.
A series of consecutive snapshots comparing the TTR heuristic with the OBS and BRT pruners are presented in Fig.~\ref{exp:compa1}.
These snapshots reveal that the BRT pruner proactively removes nodes that would drive the robot too close to obstacles, enabling the planner to adjust the robot's direction early and maintain a feasible corridor around the obstacle.
In comparison, the OBS pruner, which lacks the predictive capability, continues to expand nodes that lead to inevitable obstructions, ultimately resulting in the robot becoming stuck in a dead end.

These results underscore the critical importance of a sophisticated heuristic-pruner pairing for reliable navigation in dense, cluttered environments.

\subsection{Navigation in Dynamic Environments}
\begin{figure}[htp]
  \centering
  \includegraphics[width=0.48\textwidth]{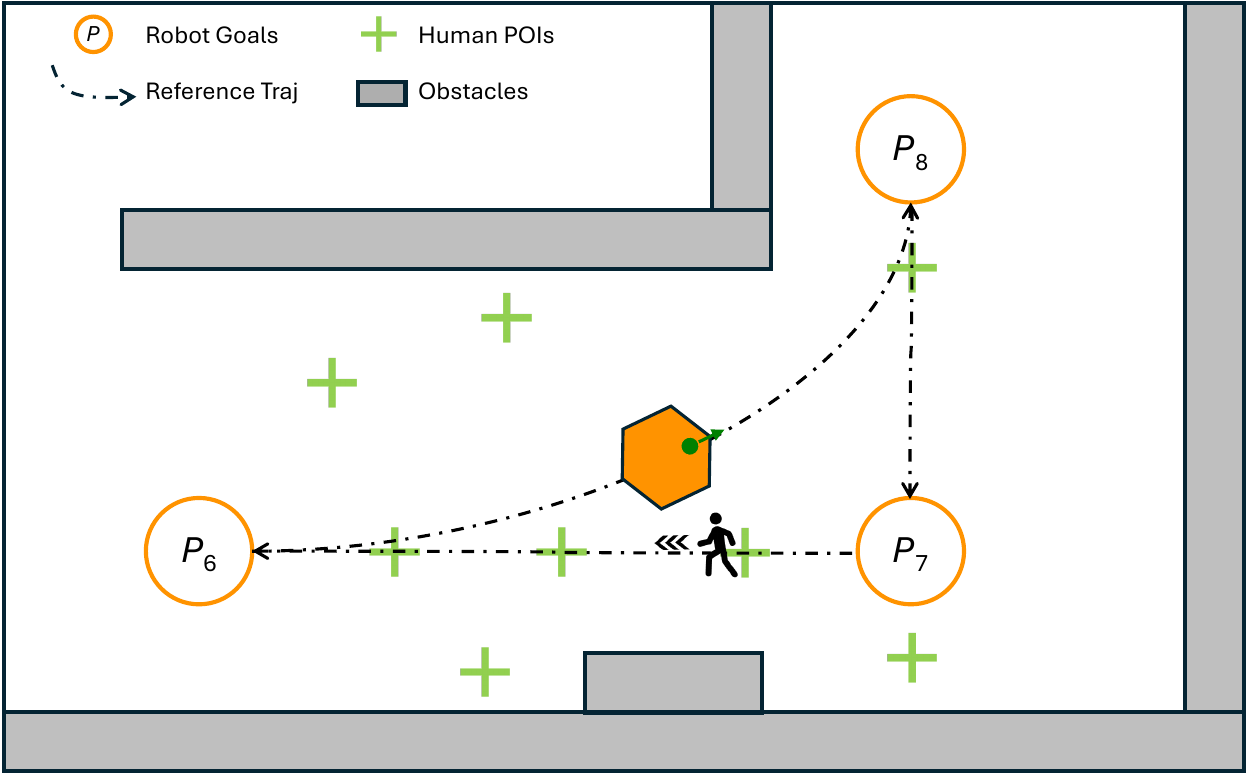}
  \caption{Layout of the dynamic environment. Dotted lines represent the reference trajectories for the robot connecting the goals ($P_6$-$P_8$). Note that humans navigate among the orange crosses, creating dynamic interference for the robot.}
  \label{dynamic-maps}
\end{figure}

Furthermore, we evaluated our method in a dynamic environment populated by a moving human pedestrian. 
Specifically, we defined three navigation goals for the robot: $P_6$ (0.8, 5.0), $P_7$ (0.8, -0.7), and $P_8$ (2.7, -0.7). 
The robot was tasked with executing a continuous patrol sequence: $P_6 \to P_7 \to P_8 \to P_7 \to P_6$ as shown in Fig.~\ref{dynamic-maps}.
To simulate realistic interaction, the human agent was assigned a set of specific behavioral rules: (1) navigate between pre-set human Point of Interests (POIs); (2) avoid the robot only when it is stationary, i.e., the human does not actively yield to a moving robot and behaves as a non-cooperative dynamic obstacle; and (3) switch to an alternative POI if the current POI is occupied by the robot.
We evaluated the proposed FI-BRT against a reactive baseline (OBS Pruner) and an ablated variant based on feasible-set pruning (FS-BRT), as described in Section~\ref{sim-dynamic}. 
We recorded the total navigation time and collision counts over three full patrol cycles.
The results are presented in Table~\ref{tab:exp2}.

\begin{table}[htb]
  \centering
  \caption{Navigation Performance in Dynamic Environments}
  \label{tab:exp2}
  \begin{tabular}{lcccc} 
      \toprule
      \multirow{2}{*}{Methods} & \multicolumn{2}{c}{Completion Time (s)} & \multicolumn{2}{c}{Collisions} \\
      \cmidrule(lr){2-3} \cmidrule(lr){4-5} 
       & 1 Human & 2 Humans & 1 Human & 2 Humans \\ 
      \midrule
      
      \textbf{FI-BRT}  & 157.6 & 264.3 & 0 & 0 \\
      \textbf{FS-BRT}  & 118.2 & 171.3 & 0 & 0 \\
      OBS & 111.5 & 134.9 & 1 & 3 \\
      \bottomrule
  \end{tabular}
\end{table}

In terms of safety, both the FI-BRT and FS-BRT methods outperformed the OBS baseline by maintaining a zero-collision record. 
This behavior is consistent with the predictive safety reasoning enabled by BRT-based pruning, which enables both planners to reason about future interactions with dynamic obstacles over a finite or infinite horizon. 
In contrast, the OBS planner, lacking this predictive foresight and thus unable to stop in advance, resulted in inevitable collisions in both scenarios, despite achieving the shortest navigation times.

While both BRT-based methods ensure safety, their efficiency differs significantly due to their distinct safety formulations. 
The FI-BRT method adopts a strictly "far-sighted" perspective. 
It seeks to maintain the robot within a guaranteed safe region indefinitely. 
This formulation makes the planner highly conservative: the robot frequently decelerates or stops to preserve the invariant set property when facing uncertainty, leading to the longest completion time (157.6 s and 264.3 s). 
Conversely, the FS-BRT method focuses on the pre-defined time horizon. 
While still predictive, it is less restrictive than the invariant approach which reduces stopping duration, allowing for more fluent motion and reduced travel time (118.2 s and 171.3 s), albeit with a theoretically smaller safety margin than the infinite-horizon guarantee of the FI-BRT.
These results further highlight the role of FS-BRT as an ablated variant that trades long-term safety guarantees for improved execution efficiency.
The qualitative difference in behavior, particularly the conservative stop-and-go motion of the FI-BRT method versus the fluent navigation of the FS-BRT method, is clearly demonstrated in the supplementary video.
Overall, these experiments confirm that reachability-based pruning provides a principled mechanism for balancing safety and efficiency in dynamic human-robot environments.
\section{Conclusion}

In this paper, we present an HJ reachability-guided framework designed to enhance the efficiency and safety of graph search-based methods for planar robot navigation.
By precomputing three HJ reachability value functions, including a TTR function, a static map-based BRT, and a human-robot BRT, our method integrates rigorous heuristic and safety information into the online search process.
Specifically, the TTR value function serves as a dynamics-informed heuristic to guide the search process; while the BRTs act as pruning mechanisms to effectively eliminate unsafe nodes from the search graph.

Extensive evaluations in both simulation and real-world settings demonstrate the effectiveness of the proposed framework.
In simulation, the TTR heuristic significantly improves search efficiency, while the static-map BRT enhances both safety and computational performance.
The FI human-robot BRT further enables robust collision-free navigation in dynamic environments, revealing a clear safety-efficiency trade-off relative to its finite-horizon counterpart.
Real-world experiments confirm that the framework generalizes beyond simulation and remains effective under sensing and control constraints.

A key insight of this work is the separation between offline reachability computation and online planning. 
By shifting the computational burden of solving HJ PDEs offline, the proposed framework enables real-time search to leverage rich dynamical and safety information without recomputing value functions online.
This provides a practical pathway for integrating HJ reachability analysis—traditionally studied in optimal control—into discrete search-based planning.

Despite these promising results, several limitations remain.
First, the offline computation of HJ value functions is computationally intensive, which may restrict scalability to higher-dimensional systems or large-scale environments.
Second, the pursuit-evasion formulation used for the human-robot BRT relies on a worst-case assumption, which can occasionally result in overly conservative navigation behaviors.
Future work will focus on addressing these challenges by improving the scalability of reachability computations, incorporating more expressive and adaptive human models, and extending the framework to handle increasingly complex and unstructured environments.

\appendix

\section{Implementation Essentials: FI-BRT Parameters}
\label{app:fibrt}

This appendix summarizes the parameter settings used for the FI-BRT pruning strategy.
Unless otherwise stated, the same parameters are used consistently across simulation and real-world experiments.

\subsection{FI-BRT Pruning Parameters}
\label{app:fibrt_params}

The FI-BRT pruning rule (Eq.~\eqref{FI-Prune}) relies on a set of parameters that regulate the conservativeness of the forward-invariant safety constraint.
Table~\ref{tab:fibrt_params} lists the values used in all experiments.

\begin{table}[htbp]
    \centering
    \caption{FI-BRT parameters used in both simulation and real-world experiments.}
    \label{tab:fibrt_params}
    \begin{tabular}{llp{3.6cm}}
      \toprule
      Parameter & Value & Description \\
      \midrule
      $\Delta t$ & \texttt{0.5} &
      Time interval between neighboring nodes used for finite-difference approximation. \\
  
      $\epsilon_{\text{rel}}$ & \texttt{0.0} &
      Safety margin applied to the value function. \\
  
      $\alpha_{\text{low}}$ & \texttt{0.8} &
      Lower bound of the invariance rate (more conservative). \\
  
      $\alpha_{\text{high}}$ & \texttt{1.2} &
      Upper bound of the invariance rate (less conservative). \\
  
      $k$ & \texttt{10.0} &
      Angular sensitivity coefficient in Eq.~\eqref{a_align}. \\
  
      $w_1,\, w_2$ & \texttt{0.5, 0.5} &
      Weights for angular and distance alignment terms. \\
  
      $d_{\text{thre}}$ & \texttt{0.5} &
      Distance threshold in Eq.~\eqref{d_align}. \\
      \bottomrule
    \end{tabular}
  \end{table}

\balance
\bibliographystyle{ieeetr}
\bibliography{ref}

@inproceedings{yang2013one,
  title     = {One-shot computation of reachable sets for differential games},
  author    = {Yang, Insoon and Becker-Weimann, Sabine and Bissell, Mina J and Tomlin, Claire J},
  booktitle = {Proceedings of the 16th international conference on Hybrid systems: computation and control},
  pages     = {183--192},
  year      = {2013}
}

@article{mitchell2005time,
  title     = {A time-dependent Hamilton-Jacobi formulation of reachable sets for continuous dynamic games},
  author    = {Mitchell, Ian M and Bayen, Alexandre M and Tomlin, Claire J},
  journal   = {IEEE Transactions on automatic control},
  volume    = {50},
  number    = {7},
  pages     = {947--957},
  year      = {2005},
  publisher = {IEEE}
}

@inproceedings{bansal2017hamilton,
  title        = {Hamilton-jacobi reachability: A brief overview and recent advances},
  author       = {Bansal, Somil and Chen, Mo and Herbert, Sylvia and Tomlin, Claire J},
  booktitle    = {2017 IEEE 56th Annual Conference on Decision and Control (CDC)},
  pages        = {2242--2253},
  year         = {2017},
  organization = {IEEE}
}

@article{lavalle1998rapidly,
  title     = {Rapidly-exploring random trees: A new tool for path planning},
  author    = {LaValle, Steven},
  journal   = {Research Report 9811},
  year      = {1998},
  publisher = {Department of Computer Science, Iowa State University}
}

@article{crandall1983viscosity,
  title   = {Viscosity solutions of Hamilton-Jacobi equations},
  author  = {Crandall, Michael G and Lions, Pierre-Louis},
  journal = {Transactions of the American Mathematical Society},
  volume  = {277},
  number  = {1},
  pages   = {1--42},
  year    = {1983}
}

@article{hart1968formal,
  title={A formal basis for the heuristic determination of minimum cost paths},
  author={Hart, Peter E and Nilsson, Nils J and Raphael, Bertram},
  journal={IEEE transactions on Systems Science and Cybernetics},
  volume={4},
  number={2},
  pages={100--107},
  year={1968},
  publisher={IEEE}
}

@inproceedings{stentz1994optimal,
  title={Optimal and efficient path planning for partially-known environments},
  author={Stentz, Anthony},
  booktitle={Proceedings of the 1994 IEEE international conference on robotics and automation},
  pages={3310--3317},
  year={1994},
  organization={IEEE}
}

@inproceedings{agand2022human,
  title={Human navigational intent inference with probabilistic and optimal approaches},
  author={Agand, Pedram and Taherahmadi, Mahdi and Lim, Angelica and Chen, Mo},
  booktitle={2022 International Conference on Robotics and Automation (ICRA)},
  pages={8562--8568},
  year={2022},
  organization={IEEE}
}

@inproceedings{korf2000recent,
  title={Recent progress in the design and analysis of admissible heuristic functions},
  author={Korf, Richard E},
  booktitle={International Symposium on Abstraction, Reformulation, and Approximation},
  pages={45--55},
  year={2000},
  organization={Springer}
}

@inproceedings{van2011anytime,
  title={Anytime nonparametric A},
  author={Van Den Berg, Jur and Shah, Rajat and Huang, Arthur and Goldberg, Ken},
  booktitle={Proceedings of the AAAI conference on artificial intelligence},
  volume={25},
  number={1},
  pages={105--111},
  year={2011}
}

@article{hansen2007anytime,
  title={Anytime heuristic search},
  author={Hansen, Eric A and Zhou, Rong},
  journal={Journal of Artificial Intelligence Research},
  volume={28},
  pages={267--297},
  year={2007}
}

@article{kavraki1996probabilistic,
  title={Probabilistic roadmaps for path planning in high-dimensional configuration spaces},
  author={Kavraki, Lydia E and Svestka, Petr and Latombe, J-C and Overmars, Mark H},
  journal={IEEE transactions on Robotics and Automation},
  volume={12},
  number={4},
  pages={566--580},
  year={1996},
  publisher={IEEE}
}

@article{schwenzer2021review,
  title={Review on model predictive control: An engineering perspective},
  author={Schwenzer, Max and Ay, Muzaffer and Bergs, Thomas and Abel, Dirk},
  journal={The International Journal of Advanced Manufacturing Technology},
  volume={117},
  number={5},
  pages={1327--1349},
  year={2021},
  publisher={Springer}
}

@article{zhu2021deep,
  title={Deep reinforcement learning based mobile robot navigation: A review},
  author={Zhu, Kai and Zhang, Tao},
  journal={Tsinghua Science and Technology},
  volume={26},
  number={5},
  pages={674--691},
  year={2021},
  publisher={TUP}
}

@article{fisac2018probabilistically,
  title={Probabilistically safe robot planning with confidence-based human predictions},
  author={Fisac, Jaime F and Bajcsy, Andrea and Herbert, Sylvia L and Fridovich-Keil, David and Wang, Steven and Tomlin, Claire J and Dragan, Anca D},
  journal={arXiv preprint arXiv:1806.00109},
  year={2018}
}

@inproceedings{macenski2020marathon,
  title={The marathon 2: A navigation system},
  author={Macenski, Steve and Mart{\'\i}n, Francisco and White, Ruffin and Clavero, Jonatan Gin{\'e}s},
  booktitle={2020 IEEE/RSJ International Conference on Intelligent Robots and Systems (IROS)},
  pages={2718--2725},
  year={2020},
  organization={IEEE}
}

@misc{minh2025optimize,
  doi = {10.48550/ARXIV.2204.05520},
  url = {https://arxiv.org/abs/2204.05520},
  author = {Bui, Minh and Hu, Hanyang and He, Chong and Lu, Michael and Giovanis, George and Shriraman, Arrvindh and Chen, Mo},
  title = {OptimizedDP: An Efficient, User-friendly Library For Optimal Control and Dynamic Programming},
  publisher = {arXiv},
  year = {2022}, 
  copyright = {Creative Commons Attribution 4.0 International}
}

@book{edelkamp2011heuristic,
  title={Heuristic search: theory and applications},
  author={Edelkamp, Stefan and Schr{\"o}dl, Stefan},
  year={2011},
  publisher={Elsevier}
}

@inproceedings{yonetani2021path,
  title={Path planning using neural a* search},
  author={Yonetani, Ryo and Taniai, Tatsunori and Barekatain, Mohammadamin and Nishimura, Mai and Kanezaki, Asako},
  booktitle={International conference on machine learning},
  pages={12029--12039},
  year={2021},
  organization={PMLR}
}

@inproceedings{koenig2002d,
  title={D* lite},
  author={Koenig, Sven and Likhachev, Maxim},
  booktitle={Eighteenth national conference on Artificial intelligence},
  pages={476--483},
  year={2002}
}

@inproceedings{hess2016real,
  title={Real-time loop closure in 2D LIDAR SLAM},
  author={Hess, Wolfgang and Kohler, Damon and Rapp, Holger and Andor, Daniel},
  booktitle={2016 IEEE international conference on robotics and automation (ICRA)},
  pages={1271--1278},
  year={2016},
  organization={IEEE}
}

@article{orthey2023sampling,
  title={Sampling-based motion planning: A comparative review},
  author={Orthey, Andreas and Chamzas, Constantinos and Kavraki, Lydia E},
  journal={Annual Review of Control, Robotics, and Autonomous Systems},
  volume={7},
  year={2023},
  publisher={Annual Reviews}
}

@article{chen2021fastrack,
  title={Fastrack: a modular framework for real-time motion planning and guaranteed safe tracking},
  author={Chen, Mo and Herbert, Sylvia L and Hu, Haimin and Pu, Ye and Fisac, Jaime Fernandez and Bansal, Somil and Han, SooJean and Tomlin, Claire J},
  journal={IEEE Transactions on Automatic Control},
  volume={66},
  number={12},
  pages={5861--5876},
  year={2021},
  publisher={IEEE}
}

@article{lu2025safe,
  title={Safe Learning in the Real World via Adaptive Shielding with Hamilton-Jacobi Reachability},
  author={Lu, Michael and Gosain, Jashanraj Singh and Sang, Luna and Chen, Mo},
  journal={Proceedings of Machine Learning Research vol},
  volume={283},
  pages={1--14},
  year={2025}
}

@article{nakamura2025generalizing,
  title={Generalizing safety beyond collision-avoidance via latent-space reachability analysis},
  author={Nakamura, Kensuke and Peters, Lasse and Bajcsy, Andrea},
  journal={arXiv preprint arXiv:2502.00935},
  year={2025}
}

@inproceedings{lyuttr,
  title={Ttr-based reward for reinforcement learning with implicit model priors. In 2020 IEEE},
  author={Lyu, Xubo and Chen, Mo},
  booktitle={RSJ International Conference on Intelligent Robots and Systems (IROS)},
  pages={5484--5489},
  year={2020}
}

@article{borquez2024safety,
  title={On safety and liveness filtering using hamilton-jacobi reachability analysis},
  author={Borquez, Javier and Chakraborty, Kaustav and Wang, Hao and Bansal, Somil},
  journal={IEEE Transactions on Robotics},
  year={2024},
  publisher={IEEE}
}

@manual{QuanserQBotPlatform,
  author       = {{Quanser Inc.}},
  title        = {{QBot Platform} User Manual},
  year         = {2023},
  note         = {Available at \url{https://docs.quanser.com/quarc/documentation/qbot_platform.html}},
  organization = {Quanser Inc.},
  address      = {Markham, Ontario, Canada}
}

@article{macenski2024open,
  title={Open-source, cost-aware kinematically feasible planning for mobile and surface robotics},
  author={Macenski, Steve and Booker, Matthew and Wallace, Joshua and Fischer, Tobias},
  journal={arXiv preprint arXiv:2401.13078},
  year={2024}
}

@inproceedings{parkinson2022time,
  title={Time-optimal paths for simple cars with moving obstacles in the Hamilton-Jacobi formulation},
  author={Parkinson, Christian and Ceccia, Madeline},
  booktitle={2022 American Control Conference (ACC)},
  pages={2944--2949},
  year={2022},
  organization={IEEE}
}

@article{chen2018hamilton,
  title={Hamilton--jacobi reachability: Some recent theoretical advances and applications in unmanned airspace management},
  author={Chen, Mo and Tomlin, Claire J},
  journal={Annual Review of Control, Robotics, and Autonomous Systems},
  volume={1},
  number={1},
  pages={333--358},
  year={2018},
  publisher={Annual Reviews}
}

@inproceedings{likhachev2005anytime,
  title={Anytime dynamic A*: An anytime, replanning algorithm.},
  author={Likhachev, Maxim and Ferguson, David I and Gordon, Geoffrey J and Stentz, Anthony and Thrun, Sebastian},
  booktitle={ICAPS},
  volume={5},
  pages={262--271},
  year={2005}
}

@article{pohl1970heuristic,
  title={Heuristic search viewed as path finding in a graph},
  author={Pohl, Ira},
  journal={Artificial intelligence},
  volume={1},
  number={3-4},
  pages={193--204},
  year={1970},
  publisher={Elsevier}
}

@inproceedings{pivtoraiko2011kinodynamic,
  title={Kinodynamic motion planning with state lattice motion primitives},
  author={Pivtoraiko, Mihail and Kelly, Alonzo},
  booktitle={2011 IEEE/RSJ International Conference on Intelligent Robots and Systems},
  pages={2172--2179},
  year={2011},
  organization={IEEE}
}

@inproceedings{plaku2007discrete,
  title={Discrete Search Leading Continuous Exploration for Kinodynamic Motion Planning.},
  author={Plaku, Erion and Kavraki, Lydia E and Vardi, Moshe Y},
  booktitle={Robotics: Science and Systems},
  pages={326--333},
  year={2007}
}

@inproceedings{fisac2015reach,
  title={Reach-avoid problems with time-varying dynamics, targets and constraints},
  author={Fisac, Jaime F and Chen, Mo and Tomlin, Claire J and Sastry, S Shankar},
  booktitle={Proceedings of the 18th international conference on hybrid systems: computation and control},
  pages={11--20},
  year={2015}
}

@article{hsu2023safety,
  title={The safety filter: A unified view of safety-critical control in autonomous systems},
  author={Hsu, Kai-Chieh and Hu, Haimin and Fisac, Jaime F},
  journal={Annual Review of Control, Robotics, and Autonomous Systems},
  volume={7},
  year={2023},
  publisher={Annual Reviews}
}

@article{chen2018robust,
  title={Robust sequential trajectory planning under disturbances and adversarial intruder},
  author={Chen, Mo and Bansal, Somil and Fisac, Jaime F and Tomlin, Claire J},
  journal={IEEE Transactions on Control Systems Technology},
  volume={27},
  number={4},
  pages={1566--1582},
  year={2018},
  publisher={IEEE}
}

@inproceedings{ferber2022neural,
  title={Neural network heuristic functions for classical planning: Bootstrapping and comparison to other methods},
  author={Ferber, Patrick and Gei{\ss}er, Florian and Trevizan, Felipe and Helmert, Malte and Hoffmann, J{\"o}rg},
  booktitle={Proceedings of the International Conference on Automated Planning and Scheduling},
  volume={32},
  pages={583--587},
  year={2022}
}

@article{savinov2018semi,
  title={Semi-parametric topological memory for navigation},
  author={Savinov, Nikolay and Dosovitskiy, Alexey and Koltun, Vladlen},
  journal={arXiv preprint arXiv:1803.00653},
  year={2018}
}

@inproceedings{guez2018learning,
  title={Learning to search with mctsnets},
  author={Guez, Arthur and Weber, Th{\'e}ophane and Antonoglou, Ioannis and Simonyan, Karen and Vinyals, Oriol and Wierstra, Daan and Munos, R{\'e}mi and Silver, David},
  booktitle={International conference on machine learning},
  pages={1822--1831},
  year={2018},
  organization={PMLR}
}

@article{parthasarathy2023c,
  title={C-MCTS: Safe planning with Monte Carlo tree search},
  author={Parthasarathy, Dinesh and Kontes, Georgios and Plinge, Axel and Mutschler, Christopher},
  journal={arXiv preprint arXiv:2305.16209},
  year={2023}
}

@article{primatesta2019risk,
  title={A risk-aware path planning strategy for UAVs in urban environments},
  author={Primatesta, Stefano and Guglieri, Giorgio and Rizzo, Alessandro},
  journal={Journal of Intelligent \& Robotic Systems},
  volume={95},
  number={2},
  pages={629--643},
  year={2019},
  publisher={Springer}
}

@article{yan2023safe,
  title={A safe heuristic path-planning method based on a search strategy},
  author={Yan, Xiaozhen and Zhou, Xinyue and Luo, Qinghua},
  journal={Sensors},
  volume={24},
  number={1},
  pages={101},
  year={2023},
  publisher={MDPI}
}

@article{karaman2011sampling,
  title={Sampling-based algorithms for optimal motion planning},
  author={Karaman, Sertac and Frazzoli, Emilio},
  journal={The international journal of robotics research},
  volume={30},
  number={7},
  pages={846--894},
  year={2011},
  publisher={Sage Publications Sage UK: London, England}
}

@inproceedings{jamgochian2024constrained,
  title={Constrained hierarchical monte carlo belief-state planning},
  author={Jamgochian, Arec and Buurmeijer, Hugo and Wray, Kyle H and Corso, Anthony and Kochenderfer, Mykel J},
  booktitle={2024 IEEE International Conference on Robotics and Automation (ICRA)},
  pages={2368--2374},
  year={2024},
  organization={IEEE}
}

@article{parimi2025risk,
  title={Risk-Bounded Multi-Agent Visual Navigation via Iterative Risk Allocation},
  author={Parimi, Viraj and Williams, Brian C},
  journal={arXiv preprint arXiv:2509.08157},
  year={2025}
}

@article{seo2025uncertainty,
  title={Uncertainty-aware Latent Safety Filters for Avoiding Out-of-Distribution Failures},
  author={Seo, Junwon and Nakamura, Kensuke and Bajcsy, Andrea},
  journal={arXiv preprint arXiv:2505.00779},
  year={2025}
}

@article{janson2015fast,
  title={Fast marching tree: A fast marching sampling-based method for optimal motion planning in many dimensions},
  author={Janson, Lucas and Schmerling, Edward and Clark, Ashley and Pavone, Marco},
  journal={The International journal of robotics research},
  volume={34},
  number={7},
  pages={883--921},
  year={2015},
  publisher={SAGE Publications Sage UK: London, England}
}

@inproceedings{ames2019control,
  title={Control barrier functions: Theory and applications},
  author={Ames, Aaron D and Coogan, Samuel and Egerstedt, Magnus and Notomista, Gennaro and Sreenath, Koushil and Tabuada, Paulo},
  booktitle={2019 18th European control conference (ECC)},
  pages={3420--3431},
  year={2019},
  organization={Ieee}
}

\end{document}